\documentclass[11pt]{article}
\usepackage[margin=1in]{geometry}
\usepackage{amsmath,amssymb,amsthm}
\usepackage{hyperref}
\hypersetup{pdftitle={Algorithmic Information Dynamics of Learning: A Certified,
 Differentiable Complexity Controller for Grokking},
 pdfauthor={Hector Zenil and Luan Ozelim}}
\usepackage{booktabs}
\usepackage{graphicx}
\usepackage{authblk}
\graphicspath{{.}{figures/}}

\usepackage[section]{placeins}

\theoremstyle{definition}
\newtheorem{condition}{Condition}

\newcommand{\sF}{{\mathrm{s}F}}
\newcommand{\KSol}{K^{\mathrm{CDM}}_{\sF}}
\newcommand{\KSolsoft}{K^{\mathrm{soft}}_{\sF}}  

\title{Algorithmic Information Dynamics of Learning:\\
A Certified, Differentiable Complexity Controller for Grokking}
\author[1]{Luan Ozelim}
\author[1]{Abicumaran Uthamacumaran}
\author[1,2]{Hector Zenil\thanks{Corresponding author: hector.zenil@kcl.ac.uk}}

\affil[1]{\normalsize\text{ }Oxford Immune Algorithmics, Oxford University Innovation \& London Institute for Healthcare Engineering, U.K}
\affil[2]{\normalsize\text{ }Department of Biomedical Computing, School of Biomedical Engineering and Imaging Sciences \& King's Institute for AI, King's College London, U.K}

\date{}
\begin{document}
\maketitle

\begin{abstract}
Algorithmic Information Dynamics (AID) studies systems by perturbing them and
measuring changes in algorithmic complexity, but its usual estimator, the
Block Decomposition Method, is piecewise constant, restricting the calculus to
finite differences.  We use $K^{\mathrm{CDM}}_{\mathrm{s}F}$, a certified, \emph{differentiable}
estimator, to bring the calculus into learning dynamics: grokking,
where a complexity \emph{order parameter} is known
but has not been made to \emph{act}.  As a transient loss kick, the estimator
becomes a \emph{controller} that accelerates grokking in Levin's
description-length--versus-time sense, within a data-dependent Occam
boundary whose finite-size trend, $f_c\sim\ln p/p$, is consistent
with a coupon-collector interpretation.  Ablations show that a complexity gate
matches a train-loss gate in rescuing failing seeds with $27\%$ less
intervention; among the tested signals, only map complexity marks the transition's
\emph{completion}; the certified prior and the per-parameter $\nabla K$
attribution are both fungible (a uniform-prior sensor makes bit-identical gate
decisions, and random supports match $\nabla K$-selected ones above a sparsity
threshold); and direct field perturbation shows a \emph{nucleation-like}
response to the Occam field (no linear regime is resolved over the probed
amplitudes, so these measurements do not justify a
fluctuation--dissipation surrogate), with a finite-field response growing by
orders of magnitude toward the phase-transititon.  These measurements account for
the empirically tuned staircase: bang--bang pulses, stall-fired and released
on yield, whose iteration plausibly builds the response it exploits.  The
kick transfers to sparse parity and to a transformer; a
sustained weight-space loss fails.  The algorithmic estimator's distinct
contribution is \emph{timing} (when to fire and when to release), not
attribution.
\end{abstract}

\vspace{0.6em}
\noindent\textbf{Highlights}
\begin{itemize}\itemsep1pt \parskip0pt
\item A certified, differentiable complexity estimator is used to \emph{control}, not
      only describe, grokking.
\item Delivered as a transient kick, it roughly doubles grokking speed and
      rescues failing seeds.
\item Acceleration is bounded by a data-dependent Occam boundary whose
      measured threshold recedes approximately as $\ln p/p$.
\item The response to the Occam field is nucleation-like, with no linear
      regime resolved over the probed amplitudes.
\item The complexity gate matches a loss gate using $27\%$ less intervention.
\end{itemize}

\vspace{0.4em}
\noindent\textbf{Keywords:} algorithmic information dynamics; algorithmic
complexity; grokking; control of learning dynamics; order parameter;
finite-size scaling; nucleation

\section{Introduction}
\label{sec:intro}

Algorithmic Information Dynamics (AID)
\cite{zenil2019aid,zenil2018aidbook} studies a system through the response of its
algorithmic (Kolmogorov) complexity \cite{kolmogorov1965,livitanyi2019} to
perturbation: for an element $e$ of an object $G$, the
signed quantity $\Delta K(e)=K(G\setminus e)-K(G)$ separates elements that inject
algorithmic randomness (their removal lowers complexity) from elements that
carry the object's program (their removal raises it).  Steering a system by
acting selectively on these elements is AID's \emph{algorithmic causal calculus},
and guiding machine learning by algorithmic probability is its natural
extension \cite{hernandez2021apml}.

Two things have limited the calculus.  First, its estimator: the Block
Decomposition Method \cite{zenil2018bdm}, built on the Coding Theorem Method's
enumeration of small machines \cite{delahaye2012ctm,soler2014ctm}, composes
block complexities under a block-independence
assumption and is piecewise constant in its input, so $\Delta K$ must be obtained by
finite differences over a combinatorial set of perturbations, and no gradient
exists.  Second, its domain: the calculus has been applied to \emph{objects}
(strings, networks), not to the \emph{dynamics} that produce them.

This paper addresses both limits, in the setting where the connection between
algorithmic complexity and learning dynamics is sharpest: grokking, the delayed
transition from memorisation to generalisation \cite{power2022grokking}, whose
mechanism has been studied through circuit efficiency \cite{varma2023circuit},
representation-learning phases \cite{liu2022effective}, optimiser dynamics
\cite{thilak2022slingshot}, and mechanistic progress measures
\cite{nanda2023progress}.  That a
complexity measure tracks this transition is established: DeMoss et
al.~\cite{demoss2024complexity} show a compression-based complexity of the
weights rises and falls across it; Sakabe et
al.~\cite{sakabe2026binarized} show, with the block decomposition method, that
binarised networks move toward algorithmic simplicity over training and that
this tracks the loss more closely than entropy does; and the companion paper \cite{cdm2026}
shows the certified estimator $\KSol$, read label-free from the network's own
input--output map, is a clean order parameter for it.  The \emph{descriptor} is
therefore prior art, and we treat it as our starting point.

The claim of this paper is the \emph{controller}.  We take $\KSol$, the certified,
differentiable, multidimensional estimator of \cite{cdm2026}, whose relaxation
to the real-valued hypercube is exact on the binary corners; because it is
differentiable, AID's perturbation calculus becomes a gradient, $\Delta K \to
\nabla K$.  We then use the measure to act on the training trajectory: to
accelerate the transition (as a transient kick), to time other interventions
(as a gate with a release), and to delimit when such control can work at all
(the Occam boundary and its finite-size scaling).  Each constructive result is paired with an ablation that
asks whether the algorithmic apparatus is load-bearing, and we give the
resulting negative answers the same weight as the positive ones.

\paragraph{Scope of the claim.}  Classical AID perturbs the \emph{state};
the controller below perturbs the \emph{flow} (the loss).  These are not the same
intervention, and the bridge (that in grokking the object of interest is the
network's own input--output map, which the trajectory moves through function
space) is an extension of AID rather than a direct application.  We state it as
such.

\paragraph{Roadmap.}
Section~\ref{sec:controller} builds the transient-kick controller, states the
condition under which complexity control can recover a function at all
(Condition~\ref{cond:occam}), and studies that condition's dependence on the
amount of data and on system size.  Section~\ref{sec:ablation} dissects the
controller by gating known accelerators on the order parameter and
ablating each ingredient in turn (the gate, the release, and the sensor's
certified prior), and tests whether the controller survives a change of task.
Section~\ref{sec:calculus} uses the calculus itself: it applies $\nabla K$ as a
per-parameter classification, measures the system's response to a complexity
field directly, and asks whether the measured response accounts for the
schedule that was tuned by hand.  Sections~\ref{sec:discussion}
and~\ref{sec:limits} weigh what certification buys against what the ablations
found dispensable, and set out what remains open.

\section{Complexity as a controller}
\label{sec:controller}

We train a two-layer MLP on modular arithmetic, the canonical grokking task
\cite{power2022grokking}, and read $\KSol$ of the learned input--output map (the
$p\times p$ output table) from the network's own predictions, using no
labels.  That this reading is a clean, label-free \emph{order parameter} for the
memorisation-to-generalisation transition is established in the companion paper
\cite{cdm2026}; here we ask what happens when the order parameter is used to
act.  Following \cite{cdm2026} the estimator is built on the symmetrised
reference machine $\sF$, so that $\KSol$ is exactly invariant to the arbitrary
labelling of the two output-bit values.  This choice does not affect the control
results below: because the object scored is a large, structured $p\times p$ map,
the raw and symmetrised estimates of every map that appears (true, memorised, random)
differ by under $4\%$ and induce the same ordering.  We adopt it for consistency
and because a controller read from a tape-dependent convention would be suspect.
On the raw machine $F$ the corresponding numbers are, in each case, within a few
percent of those reported.

\paragraph{The estimator, in brief.}  For a binary field (here, each of the
$\lceil\log_2 p\rceil$ bit-planes of the $p\times p$ map) $\KSol$ codes the
cells in raster order by a chain rule.  Each cell's bit is predicted by a
Bayesian mixture of two experts: a \emph{causal-context expert}, a count table
over the pattern formed by the cell's four causal neighbours, seeded with unit
pseudocount by the $\sF$ pattern law distilled from the exhaustive enumeration
of the reference machine; and a Krichevsky--Trofimov frequency expert.  The
estimate is the total codelength $\sum_i -\log_2 Q(b_i\mid b_{<i})$ in bits,
summed over bit-planes.  It is \emph{certified} in the sense of
\cite{cdm2026}: the value is an exact prefix-codelength (so
$2^{-\KSol}$ is a semimeasure), and it is exactly invariant under global bit
complementation by construction of $\sF$.  The differentiable relaxation
$\KSolsoft$ evaluates the same code on the expected bit-planes under
the network's softmax output; it coincides with $\KSol$ on binary corners and
is differentiable in the logits; every gradient of complexity used in this
paper is a gradient of $\KSolsoft$.  For brevity, references below to
``$\KSol$ pressure'' or a ``$\KSol$ loss'' mean pressure applied through this
differentiable relaxation; reported map complexities are discrete $\KSol$
values unless stated otherwise.

\subsection{Experimental protocol and reporting conventions}
Unless otherwise noted, the primary experiments use modular addition at
$p{=}31$ with a seeded random $40\%$ of the $p^2$ input pairs for training and
the complement for testing.  The model is a two-layer $\mathrm{ReLU}$ MLP with
width-$128$ operand embeddings, hidden width $256$, and a linear readout to
$p$ classes.  We train with full-batch AdamW at learning rate $10^{-3}$ and
weight decay $1.0$.  The discrete $\KSol$ readout is evaluated on the
network's complete argmax input--output map every $250$ steps; the gated
ablation uses $50$-step checks.  A run is said to grok at the first check for
which held-out accuracy exceeds $0.9$.

All reported grok-step means are conditional on reaching that criterion within
the stated budget, so every mean is accompanied by the number of successful
seeds.  Comparisons use matched seed pools within a shared code path unless
explicitly stated otherwise.  These small-sample summaries are descriptive,
not population-level estimates; per-seed results for the two principal
ablation grids and full controller constants appear in
Appendix~\ref{app:setup}.  Task-, architecture-, and actuator-specific
departures from this protocol are given where each experiment is introduced.

\paragraph{Matched follow-up analyses.}
To test whether the perturbational signal belongs to the represented object, the
parameter state, or the learning flow, we added same-snapshot counterfactual
branches at the same primary setting ($p{=}31$, training fraction $0.4$).
Twelve seeded trajectories were followed through memorisation, a snapshot
$2{,}000$ steps before grokking, the grokking check, and $2{,}000$ steps after
grokking when available. At each snapshot we compared direct one-cell map
perturbations, parameter resets to the corresponding initial values, and the
unchanged $300$-step Occam-field branch of \S\ref{sec:calculus}. The primary
state reset used $3\%$ of parameters, selected either uniformly or by
$|\partial\KSolsoft/\partial\theta_i|$, with random-reset doses of
$1\%$, $3\%$, and $10\%$ used to expose the displacement--response curve.
Discrete BDM was evaluated independently with the actual \texttt{pybdm}
implementation as a non-differentiable monitor; no Shannon/LZW proxy was used
and BDM was never substituted for the differentiable controller. Eight matched
seeds were additionally used for Hessian and representation-spectrum diagnostics,
with a shuffled-label arm as a structured-learning null. Full definitions,
uncertainty estimates, and censoring rules are given in
Appendix~\ref{app:followup}.

\subsection{From monitor to controller, and the Occam boundary}
\begin{figure}[htbp]\centering
\includegraphics[width=0.62\textwidth]{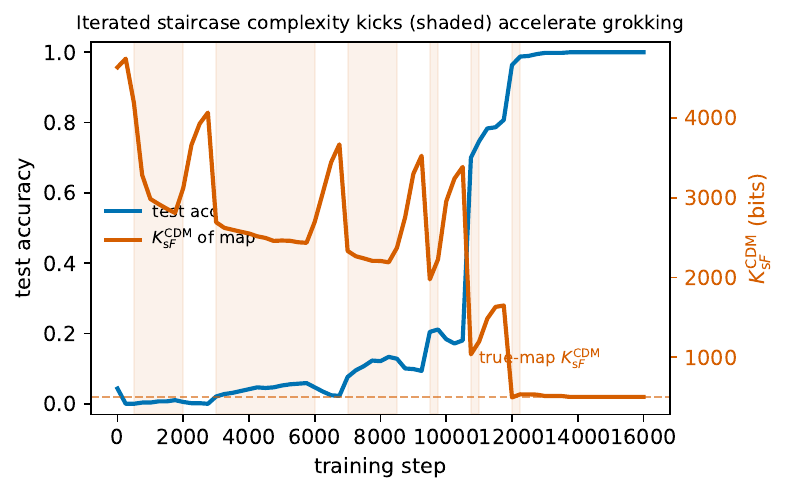}
\caption{Iterated \emph{transient} complexity kicks (shaded) accelerate grokking.
Each kick lowers $\KSol$ of the map; the network partially relapses while coasting
(the bounces), and the controller re-fires when complexity stalls.  After a few
such steps a kick breaks the memorisation basin: test accuracy jumps and $\KSol$
collapses to the true-map value (dashed).  A \emph{sustained} $\KSol$ loss instead
collapses or freezes (text); only the transient pulse drives the transition.
One representative seed; multi-seed statistics appear in the text and
\S\ref{sec:calculus}, and per-seed values for the two ablation grids in
Appendix~\ref{app:setup}.}
\label{fig:grokkick}
\end{figure}

The order parameter is also actionable, but only in a specific form and
only for a specific class of tasks.  As a \emph{sustained} loss, $\KSolsoft$ fails:
minimising it traps training in a simple but incorrect solution.  Once the data
are fit, the cross-entropy gradient vanishes and the complexity term drags the
function toward simpler maps, of which a constant map is the global minimum.
Pinning the fitted train logits to prevent this instead freezes the network:
train and test inputs share weights, and grokking is a global reorganisation
that transiently perturbs the train logits.
This is why a purpose-built
weight measure such as spectral entropy \cite{demoss2024complexity} is used as a
training loss instead.  A \emph{transient} kick escapes this: a short pulse of
$\KSol$ pressure, removed before it can change the fixed point, breaks the
memorisation basin, after which plain dynamics coast to the low-complexity
solution.  Iterating short, self-releasing kicks (each released after a modest
relative fall in $\KSol$ or a step cap, and re-fired when complexity stalls above
the best achieved) forms a push--relax--push staircase
(Fig.~\ref{fig:grokkick}).  It approximately halves the time to grok: on modular
\emph{addition} the staircase reaches full generalisation after a mean of $7{,}719$ vs.\
$17{,}833$ steps at $p{=}31$ (means over successful runs; kick $8/8$ seeds,
plain $6/8$) and $5{,}125$ vs
$9{,}375$ at $p{=}41$ ($6/6$).  It lands on the true-map complexity every time,
and rescues the seeds plain training misses.  The three established dials act on
other quantities: weight-decay scheduling \cite{varma2023circuit}, gradient
filtering \cite{lee2024grokfast} and weight-norm control
\cite{liu2022omnigrok}.

\begin{figure}[htbp]\centering
\includegraphics[width=0.98\textwidth]{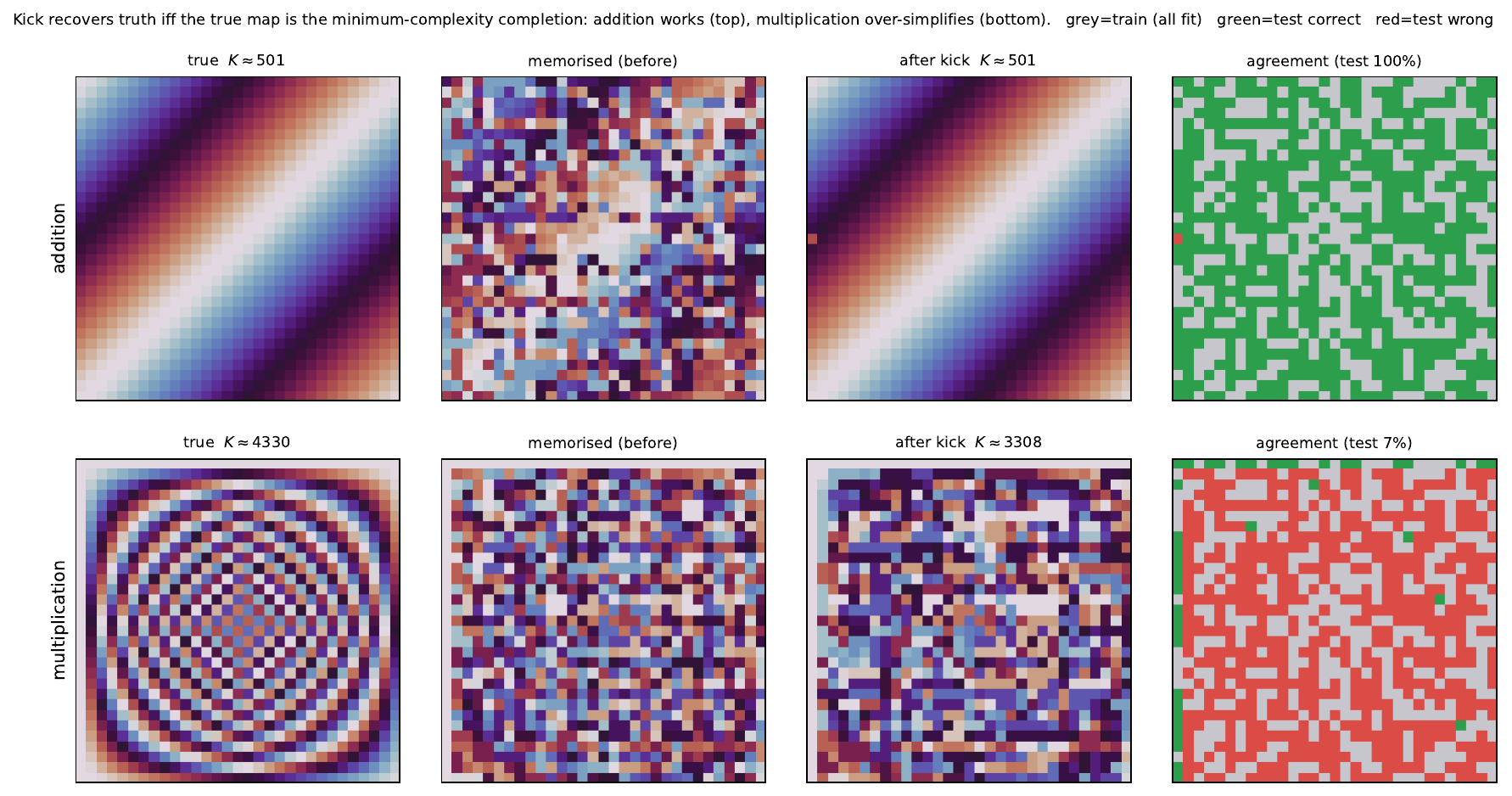}
\caption{The Occam boundary, made concrete ($p{=}31$, $40\%$ training data).
\emph{Rows:} addition (top) and multiplication (bottom).  \emph{Columns:} the true
table; the memorised map before the kick; the map after the kick; and its
agreement with truth: grey marks the $40\%$ training cells (all fit in both
cases), green/red the held-out cells that are correct/wrong.  The same
procedure recovers addition (test $100\%$; after-kick $\KSol\!\approx\!501$, the
true value) but over-simplifies multiplication to a lower-complexity patchwork
(test $7\%$; $\KSol\!\approx\!3{,}308$ against a true $4{,}330$): at this data fraction a
simpler map provides a completion consistent with the multiplication training cells
but not with the addition ones,
so the razor points at the truth for one and away from it for the other.  With
more data the multiplication gap closes (text).}
\label{fig:grokoccam}
\end{figure}

The acceleration is conditional, and the condition is Occam's razor.  The
kick presupposes that the correct solution is the minimum-complexity map
consistent with the data.  Modular addition satisfies this; modular
\emph{multiplication} does not (a lower-complexity map fits the same training
cells), and there the kick over-simplifies below the true complexity to a
wrong map ($0/6$ grok, $\KSol$ driven to ${\approx}3{,}300$ against a true-map
${\approx}4{,}330$).  This lower-$\KSol$ map is concrete
(Fig.~\ref{fig:grokoccam}): it keeps the true products on the training cells but
fills the unconstrained cells with large low-complexity patches of a few repeated
values, so it fits every training cell yet is wrong on almost all held-out
ones.  Penalising only the excess above the true value,
$\mathrm{relu}(\KSol-K^\ast)$, repairs it ($0/3\!\to\!2/3$) and leaves addition
untouched; but $K^\ast$ cannot be recovered from the training data, because
the true and the simpler-wrong map agree on every training cell, the same fact
that makes minimisation over-simplify.

The boundary is thus not merely a matter of tuning but an identifiability
constraint.

\begin{condition}[the Occam boundary]\label{cond:occam}
A necessary condition for algorithmic-complexity control to recover the true
function is that no lower-complexity function be consistent with the same
training sample.  If the true function is the unique minimum-complexity
completion and the controlled optimisation reaches that completion, complexity
pressure is aligned with recovery.
\end{condition}

\noindent Crucially, the condition is on the sample, so it
is data-dependent, and more data expands it.  Sweeping the training
fraction on multiplication, the over-simplification vanishes at a threshold
amount of data.  The kick's recovered $\KSol$ climbs $3{,}308\!\to\!3{,}605\!\to\!4{,}330\!\to\!4{,}335$
and the test accuracy jumps $0.07\!\to\!0.09\!\to\!1.00\!\to\!0.99$ as the fraction goes
$0.4\!\to\!0.5\!\to\!0.6\!\to\!0.8$.  At fraction $0.6$ the recovered complexity reaches the
true-map value ($4{,}330$) exactly as the test accuracy snaps to $1$: the true
multiplication table is then recovered at its own complexity, as predicted when
the sample makes complexity pressure point toward the true completion.

This behaviour echoes the Solomonoff principle on which the estimator is built
\cite{solomonoff1964}: with increasing data, simpler consistent hypotheses that
do not represent the generating rule are progressively excluded.
Addition's threshold is negligible (its smooth ramp is the simplest completion of
even a small sample); multiplication's is higher (its intricate structure must be
pinned by more data).  The razor accelerates learning when, and as soon as, the data make it point
at the truth.

For monitoring, the
order parameter is unconditional and label-free; for control it is
Occam-conditional, and where the condition holds the certified estimator both reads the generalisation transition
and, as a transient pulse, hastens it toward the correct solution.

\subsection{The acceleration is Levin-style, not a data--time trade}  
\begin{figure}[htbp]\centering
\includegraphics[width=0.92\textwidth]{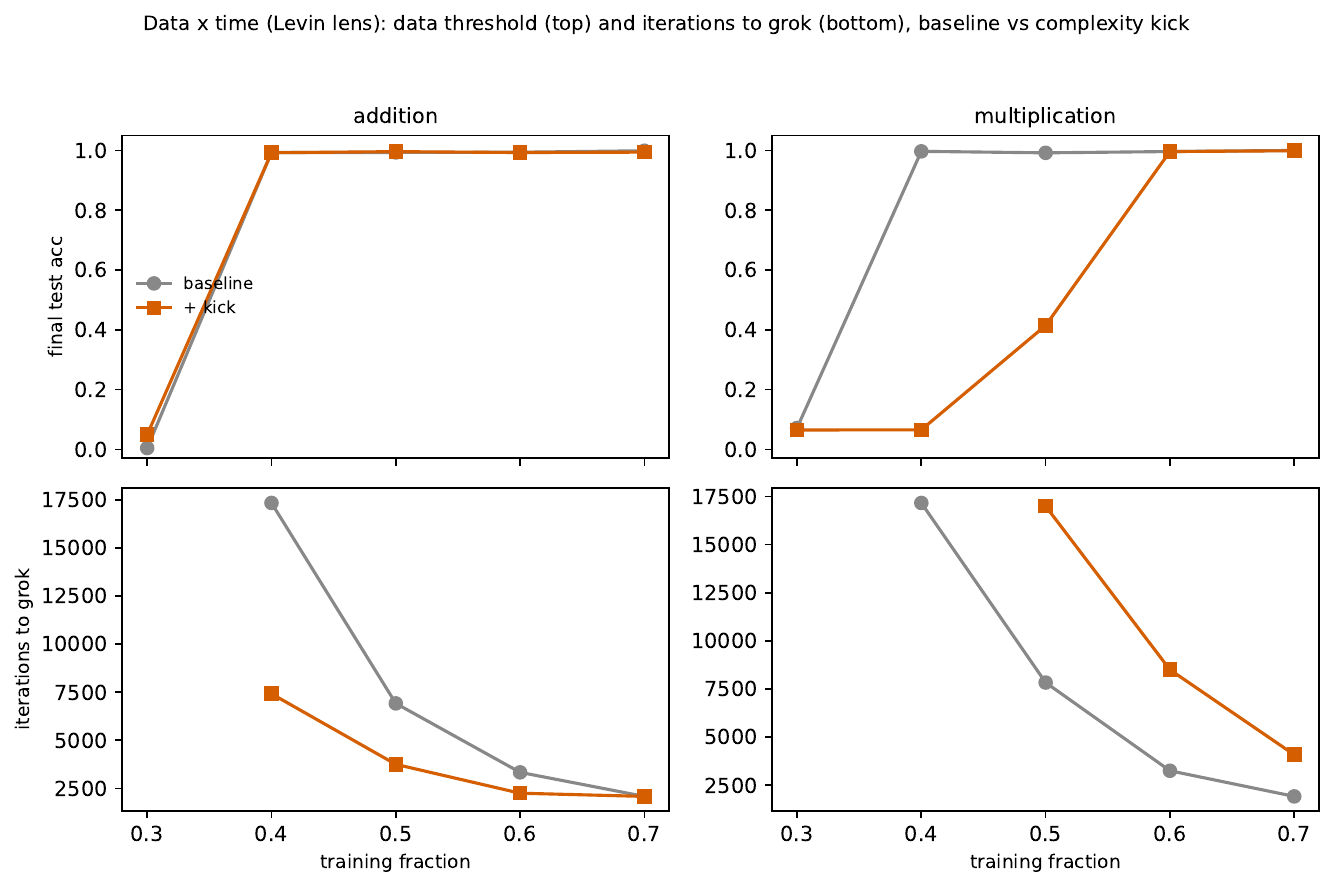}
\caption{Data\,$\times$\,time: the kick accelerates without buying speed with data.
\emph{Top:} final test accuracy vs.\ training fraction; \emph{bottom:} iterations
to grok.  On \emph{addition} (left) baseline and kick share the recovery threshold
(top curves coincide) but the kick takes fewer iterations, the gap largest at the
scarce margin ($40\%$).  On \emph{multiplication} (right) the kick recovers only
above a higher data threshold and, where both grok, takes more
iterations.  The kick is a Levin-style simplicity accelerator: faster when the
target is the minimum-complexity hypothesis, costlier in both data and time when it
is not.  Each point is a mean over three seeds; iteration means are taken over the
seeds that grok within the $30{,}000$-step budget (all three everywhere except
multiplication with the kick at $50\%$, where one of three groks); fractions at
which no seed groks are left blank.}
\label{fig:grokdatatime}
\end{figure}

It is
tempting to read the kick as buying speed with data, but a joint
data\,$\times$\,time sweep shows otherwise (Fig.~\ref{fig:grokdatatime}).  On
modular \emph{addition} the kick reaches full generalisation at the same
data threshold as plain training but in fewer iterations, and the saving is
largest at the data-scarce margin ($2.3\times$ at $40\%$ training data,
$17{,}333$ vs.\ $7{,}417$ steps, shrinking to a tie by $70\%$): same data, less
time.  On \emph{multiplication},
where Occam is misaligned below the threshold fraction, the kick is worse on
both axes: it needs more data to recover at all (threshold ${\approx}0.6$
vs.\ ${\approx}0.4$; at $50\%$ the baseline groks in all three seeds and the kick
in one) and, where both eventually grok, it is slower ($8{,}500$
vs.\ $3{,}250$ steps at $60\%$).  All figures here are means over three seeds.
So the kick is not a resource exchange; it is a
Levin-style simplicity-biased accelerator that shortens the search for the
short-program solution.  This is the operational content of Levin's
$Kt=\ell+\log t$ \cite{levin1973}: biasing the dynamics toward low description length shrinks the
time to find the low-complexity solution, a free time-saving when the
target is the minimum-complexity hypothesis the data support, and a penalty
in both time and data when it is not.  The intrinsic grokking data--time trade
(more data $\to$ shorter delay, visible in both baselines) belongs to grokking;
the kick's role is to slide the time axis down for Occam-aligned tasks.

\subsection{Finite-size scaling of the Occam boundary}
\begin{figure}[htbp]\centering
\includegraphics[width=0.98\textwidth]{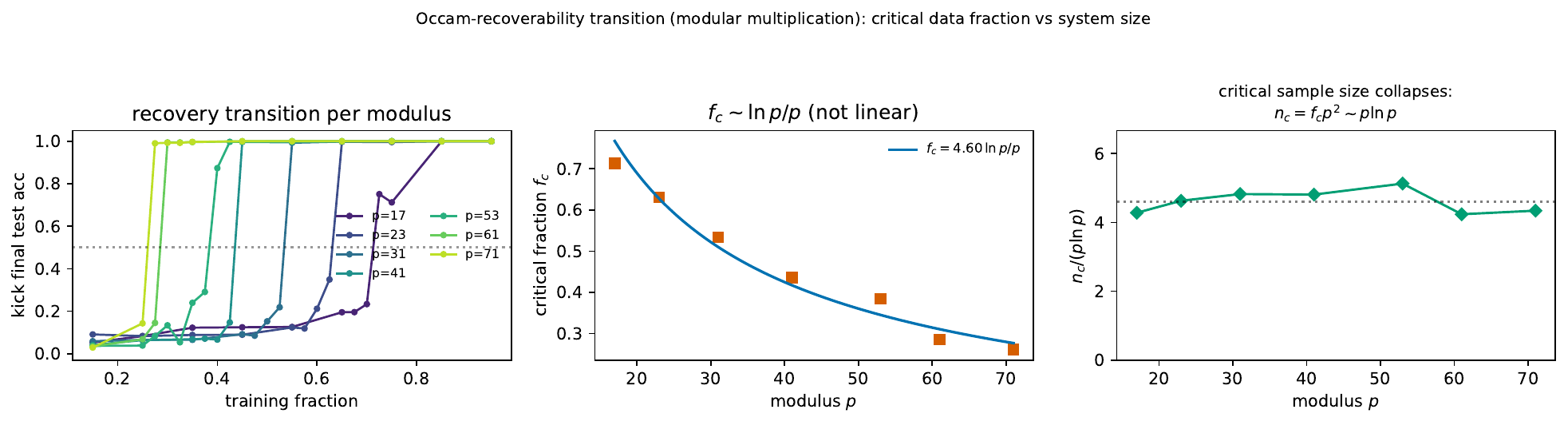}
\caption{Empirical finite-size scaling of the Occam boundary (modular multiplication), over
seven moduli $p\in\{17,\dots,71\}$.  \emph{Left:} kick recovery accuracy vs.\ training
fraction; the transition shifts left and sharpens as $p$ grows (each mean is a
$2$-seed mean, refined to $12$ seeds across the transition region, with the
recovered plateau above it filled by monotonicity).  \emph{Middle:}
the threshold fraction $f_c$ (interpolated $0.5$ crossing) against $p$, with the empirical fit
$f_c=4.6\,\ln p/p$, not a straight line to zero.  \emph{Right:} the diagnostic
collapse: the threshold sample count $n_c=f_c\,p^2$ divided by $p\ln p$ is flat
at $4.6\pm7\%$ across all seven moduli.  This normalisation is motivated by the
coupon-collector scale $p\ln p$ for covering $p$ residues and supports, but does
not uniquely establish, that interpretation.}
\label{fig:grokscaling}
\end{figure}

Because the boundary is a property of the sample, the recovery of the true
rule as the training fraction crosses $f_c$ forms a transition-like crossover in
a control parameter.  Its dependence on system size $p$ exhibits an empirical
finite-size-scaling pattern (Fig.~\ref{fig:grokscaling}).  Sweeping
$p\in\{17,23,31,41,53,61,71\}$ on multiplication, the threshold fraction falls
monotonically, $f_c\approx0.71,\,0.63,\,0.53,\,0.44,\,0.38,\,0.29,\,0.26$, while the
transition sharpens.  On a grid refined to steps of $0.025$ near $f_c$ at every
modulus, the window over which recovery accuracy lies strictly between $0.05$
and $0.95$ narrows monotonically with $p$: $0.60$, $0.48$, $0.38$, $0.18$,
$0.13$, $0.03$, and a single sampled fraction at $p{=}71$.  At $p{=}17$ the
accuracy climbs slowly across the whole sweep, from $0.05$ at $f{=}0.15$ to
$0.23$ at $f{=}0.70$, before rising; at $p{=}71$ it moves from $0.14$ to
$0.99$ across one $0.025$-wide step.  The systematic shift and sharpening are
consistent with finite-size scaling, although these finite systems and seed
counts do not by themselves establish a thermodynamic critical point.

Over the tested range, the decline is better described by $f_c\sim\ln p/p$
than by a linear extrapolation.  A possible sample-complexity explanation is
that recovery requires the training cells to constrain the multiplicative
structure across all $p$ residues.  Random coverage of $p$ categories has the
coupon-collector scale $p\ln p$, motivating this scaling ansatz rather than
deriving it from the present experiments.  The number of training constraints
is $n=f\,p^2$, so writing the threshold count as $n_c=f_c\,p^2$ and dividing by
$p\ln p$ normalises all seven moduli to a
constant, $n_c/(p\ln p)=4.6\pm 7\%$ (Fig.~\ref{fig:grokscaling}, right), flatter than the collapse under a fixed absolute sample ($n_c=\mathrm{const}$, spread
$52\%$), a fixed fraction ($n_c\propto p^2$, $34\%$), or a fixed count per residue
($n_c\propto p$, $15\%$).  These observations support
$f_c=n_c/p^2\sim\ln p/p$ within the measured range; under that ansatz the
threshold fraction vanishes only asymptotically.  (A
straight line through the first three sizes appears to reach $f_c=0$ near
$p\approx73$.  Over seven sizes that reading is an artefact of fitting a short
arc of $\ln p/p$, which tends to zero only as $p\to\infty$.)  The resulting
interpretation is Solomonoff-like: a fixed fraction supplies more absolute
evidence as $p$ grows, and simpler incorrect completions are excluded once a
sufficiently broad sample is seen.  Establishing coupon collection as the
mechanism, rather than one explanation of the observed scaling, remains open.

\section{Dissecting the controller}
\label{sec:ablation}

The kick is our own actuator.  To separate what the complexity signal
contributes from what any intervention would contribute, we now gate
known accelerators on the order parameter and ablate every ingredient:
the gate, the release, and the sensor's certified prior.  In AID terms this is
an element classification applied to the training procedure itself: the
policies below range from a blind, always-on intervention through a gate on a
cheap non-algorithmic signal (the train loss) to a gate on the algorithmic
response, and the experiment asks which classification of when to act
carries the effect.

\subsection{Gating a known accelerator on the order parameter}

\paragraph{Setup.}  Two actuators on modular addition ($p{=}31$, training
fraction $0.4$, full-batch AdamW, $25{,}000$-step budget, six seeds):
Grokfast slow-gradient amplification \cite{lee2024grokfast}
($\hat g = g + \lambda\,\mathrm{EMA}_\alpha(g)$, $\alpha{=}0.9$,
$\lambda{=}2$), and a weight-decay schedule in the spirit of the
circuit-efficiency account \cite{varma2023circuit} (decay $0.1$ normally,
boosted to $1.0$ while the gate is active).  Five gate policies each:
\emph{none} (never), \emph{always}, \emph{fixed} (from step $500$;
Grokfast's own two-stage variant), \emph{loss} (once the train cross-entropy
falls below $0.05$; a gate any cheap signal can implement, with no release),
and \emph{$\KSol$} (on once fit, released when the map's $\KSol$ has
fallen to $40\%$ of its post-memorisation peak).  Table~\ref{tab:gated} reports
the outcome; the code path is shared, and indeed the blind-on weight-decay row
reproduces the plain baseline seed for seed, as it must (both are AdamW at
decay $1.0$).

\begin{table}[htbp]\centering
\small
% generated by codes/aux/gen_ablation_table.py -- do not edit by hand
\begin{tabular}{llcccc}
\toprule
actuator & gate policy & grokked & grok step & final test & intervention steps \\
\midrule
Grokfast & \emph{none} (blind off) & 4/6 & 17,925 & 0.832 & 0 \\
 & \emph{always} (blind on) & 6/6 & 19,133 & 0.982 & 25,001 \\
 & \emph{fixed} step $\ge$ 500 & 5/6 & 18,510 & 0.901 & 24,501 \\
 & \emph{loss} (on once fit) & 6/6 & 18,833 & 1.000 & 24,884 \\
 & $\KSol$ (on once fit, released) & 6/6 & 18,766 & 0.986 & 18,208 \\
 & \quad$\KSol$, uniform-prior sensor & 6/6 & 18,766 & 0.986 & 18,208 \\
\midrule
weight decay & \emph{none} (blind off) & 0/6 & -- & 0.000 & 0 \\
 & \emph{always} (blind on) & 4/6 & 17,925 & 0.832 & 25,001 \\
 & \emph{fixed} step $\ge$ 500 & 5/6 & 17,790 & 0.928 & 24,501 \\
 & \emph{loss} (on once fit) & 6/6 & 17,858 & 0.997 & 24,884 \\
 & $\KSol$ (on once fit, released) & 3/6 & 18,633 & 0.831 & 17,500 \\
\bottomrule
\end{tabular}

\caption{Gated-vs-blind ablation ($p{=}31$, fraction $0.4$, six seeds,
$25{,}000$-step budget).  ``Grokked'' counts seeds reaching test accuracy
$>0.9$; ``grok step'' is the mean over those seeds; ``intervention steps''
is the mean number of steps the actuator was active.  The $\KSol$ gate is also
run with the sensor's certified prior replaced by a uniform base measure
(indented row): its decisions are bit-identical on every seed.}
\label{tab:gated}
\end{table}

\paragraph{The actuators buy reliability, not speed.}  In this regime
Grokfast actuation buys reliability, while the choice of gate determines how
long the intervention remains active.  Plain training groks only $4/6$ seeds
(the two failures sit at test ${\approx}0.5$ for the whole budget); the
always-on, loss-gated, and $\KSol$-gated Grokfast policies rescue all six seeds
(the fixed-step gate rescues five), at mean grok steps inside the baseline's own seed spread
($14{,}050$--$20{,}350$; per-seed values in Table~\ref{tab:gatedseeds}).  Grokfast's
headline accelerations arise in minibatch training, where the gradient
has genuinely fast stochastic components for the low-pass filter to separate;
in full batch the gradient is the slow component, so
$g+\lambda\,\mathrm{EMA}(g)\approx(1+\lambda)g$, a rescaling that Adam's
per-coordinate normalisation largely absorbs.  We verified this is not a
tuning accident by probing the other three cells of the
$\alpha\in\{0.9,0.98\}\times\lambda\in\{2,5\}$ grid (two seeds each; the
fourth cell is the always-on row of Table~\ref{tab:gated}).  No cell shows
systematic acceleration: probe grok steps span $13{,}750$--$21{,}400$,
straddling the plain baseline's own $14{,}050$--$20{,}350$ seed spread
rather than shifting below it.  The informative comparison is therefore not
speed but who rescues, and at what intervention budget.

\paragraph{The complexity gate matches the loss gate at $27\%$ less
intervention.}  It does so because the order parameter carries a signal the
loss cannot: completion.  The loss gate and the $\KSol$ gate both rescue all six
seeds with nearly equal mean grok steps ($18{,}833$ vs $18{,}766$; final test
$1.000$ vs $0.986$).  But the train loss is pinned near zero from the moment
of memorisation onward: it can switch the actuator on, never off.  The loss
gate therefore intervenes for essentially the whole budget ($24{,}884$ steps on
average).  $\KSol$ falls to the true-map value exactly when the
reorganisation completes, and the release fires on that collapse: mean
$18{,}208$ intervention steps, $27\%$ fewer, with the release trailing the
test-accuracy crossing on every seed (by $1.9$--$3.8$k steps, mean
${\approx}2.6$k).  In the AID
classification, the loss is an \emph{onset} observable; the algorithmic
response is the only observable in the set that also marks \emph{completion}.

\subsection{The release, and which actuators tolerate it}

\paragraph{Transient gating suits flow-shaping actuators; objective-shaping
actuators must persist.}  For the weight-decay actuator the same release is
harmful: $3/6$ grokked, mean final test $0.831$: released seeds relapse,
because dropping the decay back to $0.1$ restores the memorisation
optimum rather than merely slowing the dynamics.  The loss-gated (never
released) schedule is instead the best weight-decay policy ($6/6$, mean
$17{,}858$).  This is the mirror image of the kick section's dichotomy: a
\emph{sustained} $\KSol$ loss fails where the \emph{transient} kick succeeds
because the sustained version moves the fixed point; here a \emph{transient}
decay boost fails where the \emph{sustained} one succeeds because the decay
is part of the objective.  The rule that unifies all four cells:
interventions that reshape the \emph{dynamics} (gradient filtering, the kick)
should be gated transiently and released at completion; interventions that
reshape the \emph{objective} (weight decay) must persist.  The order
parameter times both correctly: it tells the first class when to let go, and
the second class when to switch on.

\paragraph{Weight-norm control is a second objective-shaping actuator.}  The
third known dial obeys the same rule.  Omnigrok
\cite{liu2022omnigrok} identifies the weight norm as the variable controlling
the grokking delay, operating on the initialisation scale.  Turned into
an in-training actuator (project the full parameter vector back to a sphere of
radius $\rho\,\|\theta_0\|$ while the gate is active), it separates sharply by
decay regime (Table~\ref{tab:wnorm}, three seeds throughout).  At the paper's
strong decay ($1.0$) the projection is simply neutral: every radius
groks in all three seeds, and the three radius means ($16{,}500$, $20{,}633$,
$17{,}367$) bracket the plain baseline's own ($17{,}817$).  At weak decay
($0.01$), where Omnigrok's effect lives, plain training never groks and ends at
the floor ($0.001$ mean final test accuracy), and so do $\rho{=}0.5$ and
$\rho{=}0.7$.  One radius engages, and does so sharply: at $\rho{=}0.3$ all three
seeds reach $0.79$--$0.82$ final test accuracy within ${\approx}3{,}000$ steps,
two of them crossing the $0.9$ grok criterion.  The dial is real but narrow:
the response is not monotone in $\rho$, and no setting here reaches the full
generalisation the strong-decay runs attain.

That operating point makes the dichotomy's prediction testable after all
(Table~\ref{tab:wnormgated}).  A norm constraint shapes the \emph{objective},
so by the rule above its release should harm, and it does.  The $\KSol$ gate
and the train-loss gate reach the same grok step on every seed (mean
$3{,}050$), but the released version ends at $0.808$ against the never-released
$0.912$.  The two seeds on which the release actually fires make the trade
explicit: the constrained budget falls from $14{,}251$ and $13{,}900$ steps to
$1{,}700$ and $1{,}250$, factors of $8.4$ and $11.1$, and final test accuracy
falls with it, from $0.958$ and $0.964$ to $0.821$ and $0.790$.  On the third
seed $\KSol$ never collapses, the release never fires, and the two policies
coincide exactly.  Always-on control is less reliable and more expensive than
the $\KSol$-gated policy ($1/3$ vs.\ $2/3$ grokked and $25{,}001$ vs.\ $5{,}417$
constrained steps).  Weight-norm control therefore
joins weight decay as a second objective-shaping actuator rather than standing
as an exception.  The order parameter identifies a low-cost release point, but
the subsequent loss of accuracy shows that release is not the correct action
for this actuator.  As with the decay schedule, the effective policy is to
switch on and maintain the constraint.

\begin{table}[htbp]\centering
\caption{Weight-norm constraint as an in-training actuator, by decay regime.
Grok step is the mean over the seeds crossing $0.9$ test accuracy, with the
count; final test accuracy is the mean over all three seeds.  Neutral at strong
decay; at weak decay only $\rho{=}0.3$ engages.}
\label{tab:wnorm}
% generated by codes/aux/gen_wnorm_table.py -- do not edit by hand
\begin{tabular}{llcc}
\toprule
regime & constraint & grok step & final test acc \\
\midrule
strong decay ($1.0$) & none (plain) & 17,817 (3/3) & 0.991 \\
 & $\rho=0.4$ & 16,500 (3/3) & 0.985 \\
 & $\rho=0.6$ & 20,633 (3/3) & 0.968 \\
 & $\rho=0.8$ & 17,367 (3/3) & 0.977 \\
\midrule
weak decay ($0.01$) & none (plain) & -- (0/3) & 0.001 \\
 & $\rho=0.3$ & 3,050 (2/3) & 0.808 \\
 & $\rho=0.5$ & -- (0/3) & 0.002 \\
 & $\rho=0.7$ & -- (0/3) & 0.000 \\
\bottomrule
\end{tabular}

\end{table}

\begin{table}[htbp]\centering
\caption{Gated vs.\ blind weight-norm control at the one engaging operating
point ($\rho{=}0.3$, weak decay), three seeds.  The $\KSol$ gate matches the
train-loss gate's grok step at $2.6\times$ less intervention overall ($5{,}417$
vs.\ $13{,}817$ constrained steps), but releasing an objective-shaping
constraint costs final accuracy.}
\label{tab:wnormgated}
% generated by codes/aux/gen_wnorm_table.py -- do not edit by hand
\begin{tabular}{lccc}
\toprule
gate & grok step & final test acc & constrained steps \\
\midrule
none (no constraint) & -- (0/3) & 0.001 & 0 \\
always (blind, never released) & 1,850 (1/3) & 0.790 & 25,001 \\
fixed step $T{=}500$ (blind) & 2,575 (2/3) & 0.860 & 24,501 \\
train-loss gate (no release) & 3,050 (2/3) & 0.912 & 13,817 \\
$\KSol$ gate (released) & 3,050 (2/3) & 0.808 & 5,417 \\
\bottomrule
\end{tabular}

\end{table}

\subsection{Ablating the sensor's certified prior}

\paragraph{The certified prior is not load-bearing for gate timing.}  The companion paper measures the $F$ prior's
contribution to the estimator's codelength at ${\sim}0.5\%$ on its corpus, an
$O(1)$-bit effect.  A controller claim built on a ``certified estimator''
therefore owes an ablation: we reran the $\KSol$-gated Grokfast experiment
with the sensor's base measure replaced by the uniform law ($G[\cdot]=1/2$
for every causal pattern, in place of the $\sF$ pattern law), leaving the
count-based context expert otherwise untouched.  The gate's decisions are
bit-identical on all six seeds (Table~\ref{tab:gated}): every
amplification window opens and closes at the same check.  The mechanism is
plain: on the true addition map the $\sF$ base measure shifts the sensor by
$1.8$ bits out of $501$ ($0.36\%$, matching the companion measurement), and
on a random map by $0.0$ bits out of $4{,}829$; against pattern counts
accumulated over $961$ cells, a unit pseudocount of base measure cannot move
a $60\%$-collapse threshold.

What the timing signal actually uses is the
estimator's structure (the causal-context mixture that measures the
map's own self-similarity), not the enumeration-derived prior that seeds it.
The claim we can support is therefore narrower: the estimator is certified in
the stated sense (its discrete values are exact prefix-codelengths, its
relaxation agrees on binary corners, and its invariances are proved).  This
licenses reading its collapse as a reduction in description length under the
specified code; but the control-relevant information would survive
an uncertified base measure, so the controller programme stands on the
mixture construction, not on the $F$ enumeration.  This is consistent with,
and sharpens, the companion's finding that the algorithmic prior's benefit is
$O(1)$ bits: at control granularity, $O(1)$ bits is below the actuation
threshold.

\subsection{What transfers, and where the domain ends}

\begin{figure}[htbp]\centering
\includegraphics[width=0.98\textwidth]{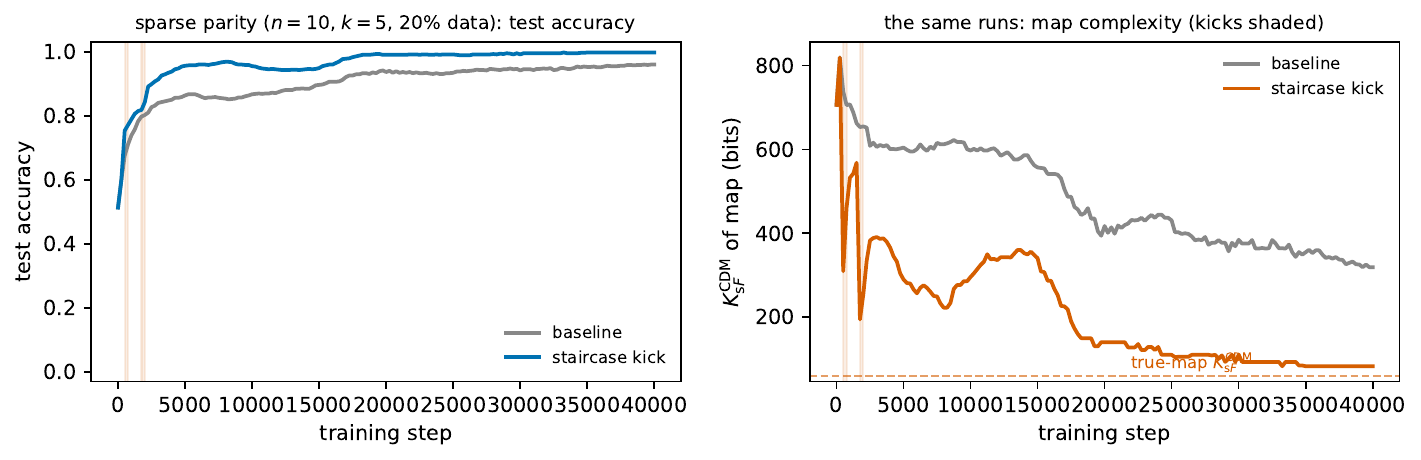}
\caption{The staircase kick on sparse parity ($n{=}10$, $k{=}5$, $20\%$
training data; the seed with the slowest baseline).  \emph{Left:} test
accuracy; the kicked run (blue) groks at $2{,}750$ steps where the baseline
(grey) crawls to its transition near $15{,}500$.  \emph{Right:} the same runs'
map complexity: the kick (orange; kick windows shaded) drives $\KSol$ down to
the true parity map's $59$ bits, while the baseline lingers at
intermediate-complexity maps throughout its pre-grok interval.  Same controller, sensor
and schedule as modular arithmetic; only the task changed.}
\label{fig:grokparity}
\end{figure}

\paragraph{The staircase kick transfers to a second task family: sparse
parity.}  Everything above is modular arithmetic, so we port the controller,
unchanged, to sparse parity, the canonical hard case for gradient learning
\cite{barak2022parity}: $n{=}10$ input bits, label the XOR of the first
$k{=}5$ (the rest distractors), an MLP trained on a fraction of the $2^{10}$
inputs, and $\KSol$ read from the network's predicted output map reshaped to a
$32\times32$ binary field.  We apply the same staircase kick (fire once fit, release on
a relative complexity fall, re-fire on a stall), with identical staircase
hyperparameters (release fraction, stall rule, kick cap, pressure schedule);
only the fit tolerance changed with the loss ($\mathrm{BCE}<0.03$ for the
single-bit output vs $\mathrm{CE}<0.01$).  At the data-scarce margin ($20\%$ of inputs,
$40{,}000$-step budget, four seeds) the baseline groks with a heavy-tailed
delay (steps $15{,}500$, $1{,}500$, $6{,}500$, $1{,}750$ across seeds), while
the kicked runs grok at $2{,}750$, $1{,}000$, $1{,}750$, $1{,}000$: every seed
accelerates, the slowest by $5.6\times$, and the mean falls from ${\sim}6{,}300$
to ${\sim}1{,}600$ steps (Fig.~\ref{fig:grokparity}).

Just above the margin
($25\%$ of inputs) the baseline delay vanishes and the kick simply matches it
($750$--$1{,}500$ vs $500$--$1{,}250$ steps), the same
largest-at-the-scarce-margin profile as modular addition's data$\times$time
sweep, and never a cost.

The complexity readout
repeats the modular-arithmetic signature: the true parity map
scores $59.4$ bits under the sensor, and the kicked runs drive the map to
$\KSol=59$ bits on three of four seeds ($82$ on the fourth), so the controller
again lands on the true-map $\KSol$ value, while the two slow baseline seeds
stall for thousands of steps at intermediate maps of $7$--$10\times$ that
complexity (median $\KSol$ over each seed's pre-grok window, $430$ and $602$
bits).  Parity is, like
addition, a task whose true rule is the minimum-complexity completion of a
modest sample (the $k{=}5$ XOR map is $59$ bits against hundreds for the
memorised patchworks), so Condition~\ref{cond:occam} predicts the kick helps, and it does.  The transfer
required changing the task, the architecture (a plain $\mathrm{ReLU}$ MLP on
bit vectors), the output geometry (a $32\times32$ single-bit field instead of
$\lceil\log_2 p\rceil$ bit-planes), and the label structure, and the
controller's behaviour is unchanged: this is evidence the mechanism is the
measure, not an artefact of modular tables.

\paragraph{And to a second architecture: a transformer.}  Returning to
modular addition at $p{=}31$, a two-block causal transformer (token embeddings for the two
operands, learned positions, $4$ heads, width $128$) replaces the MLP on the
same task, with the same map readout, staircase, and hyperparameters.  Plain training
is slower and less reliable than the MLP's.  By seed, the baseline/kick grok
steps for seeds $0$, $1$, and $2$ are $>{}30{,}000/13{,}750$,
$36{,}250/11{,}500$, and
$15{,}250/13{,}750$; the second baseline was observed in a run extended to
$40{,}000$ steps, while the first ended its $30{,}000$-step run at test $0.82$.
Every kicked run drives the map to $\KSol=501$ bits, the true-map value.  One
caveat: the first seed, rescued from a non-grokking baseline, relapses partway
after release (final test $0.76$ after touching $0.9{+}$), a coast-stability
difference between architectures that the MLP does not show; the other two
kicked seeds end at $0.99$.  The relapse is curable: adding a small
validation gate ($200$ points reserved from the original held-out pool and
excluded from the reported test set;
snapshot the best-validation state, revert and stop kicking on a $0.15$ drop)
stabilises every seed (finals $0.93$, $1.0$, $1.0$), at the cost of a later
grok on the fragile seed ($27{,}500$).  The complexity sensor remains
label-free, but this guard consumes additional labelled validation data.  The
order parameter, the kick, and the landing on the true-map $\KSol$ value
survive the change of architecture on all three transformer seeds; the
post-release coast does not, and
the guard that fixes it is the one component that needs labels beyond the
supervised training set.

\paragraph{A sustained weight-space $\KSol$ loss fails: the controller's
domain is function space.}  DeMoss et al.\ \cite{demoss2024complexity}
regularise a \emph{weight-space} complexity (spectral entropy) as a training
loss and report it induces grokking.  Running their setup with our measure
(soft $\KSol$ of the min--max--normalised embedding matrices, the only
differentiable weight readout available to us) fails at every amplitude
tried: across $\lambda\in\{10^{-3},10^{-2},10^{-1}\}$ (three seeds each) no
run grokked within the budget, and the final test accuracy sits at the
$1/31\approx0.03$ chance floor to within a few percentage points (means $0.04$,
$0.06$, $0.01$), against $1/3$ grokked and a mean final accuracy of $0.64$ for
the plain baseline in the same budget.  This is the expected outcome (the companion
paper already found min--max weight readouts to be weak-to-anti-correlated
order parameters), and we report it as the boundary it is: $\KSol$'s
controller role is in \emph{function space}, read from the network's
input--output map, where its collapse marks the transition; pushed into
weight space through a readout that is not an order parameter, the same
pressure carries no usable signal.  The comparison with spectral
entropy is thus not measure-vs-measure on common ground but domain-vs-domain:
purpose-built weight functionals can act in weight space; our map functional
acts on the representation for which its code is defined.

\section{The calculus, used}
\label{sec:calculus}

Two ingredients of AID's programme remain: use $\nabla K$ as a
per-element classification (not merely as a loss term), and explain the
controller's schedule using the system's measured response rather than tuning
alone.  This section does both, and then settles the question the measurements
raise on the way, the kick's amplitude; the answers reshape the programme in
the same direction as
\S\ref{sec:ablation}: the fine structure of the algorithmic signal is largely
fungible; its \emph{dynamics} (when the response is large, when it
saturates) is where the control information lives.

\subsection{\texorpdfstring{$\nabla K$}{Gradient of K} as a per-parameter classification}

\paragraph{The element classification is fungible above a support
threshold.}  The differentiable estimator assigns
every parameter an algorithmic response $s_i = |\partial \KSolsoft/\partial
\theta_i|$, AID's element classification applied to the network itself.  We
test whether it is load-bearing: at each kick start the parameters are scored,
and the kick's $K$-gradient is applied only through the top-$q$ fraction by
$s_i$, against a uniformly random mask of the same size
(Table~\ref{tab:gradgate}; six seeds per cell, twelve at $3\%$, same staircase as
\S\ref{sec:controller}).  Three facts emerge.  First, the kick survives
drastic sparsification: at $30\%$ support either selection reproduces the full
kick ($8{,}625$ and $9{,}541$ vs $7{,}875$ steps, $6/6$ each, against
$18{,}000$, $5/6$, unkicked).  Second, the controlling variable is the
size of the support, not the selection: at $10\%$ the $\nabla K$ and
random masks are indistinguishable ($12{,}166$ vs $12{,}208$).  The
classification earns a visible margin only in the sparse regime, and there its
value is chiefly reliability: extending the $3\%$ cells to twelve
seeds, the $\nabla K$ mask groks $11/12$ against random's $8/12$, with a
modest speed edge on top (faster in $6/7$ seed pairs where both grok, by $1{,}536$ steps on average); at $1\%$ both are weak.

Third, the classification is coarse and unstable.  It is coarse in that the top-$1\%$ set
lies almost entirely in one module: $77\%$ of its mass falls in the final
readout matrix (mean over $17$ kicks across six seeds), so at parameter
granularity it says little beyond ``the readout layer''.  It is unstable in
that consecutive kicks select nearly disjoint sets: Jaccard overlap
${\sim}5\%$ on average, $1.5$--$8.2\%$ across the $11$ consecutive pairs.  The
map's complexity is a function-space
quantity; pulled back to parameter space it attributes broadly and
transiently, and almost any sufficiently wide channel transmits the pressure.
This is the parameter-space twin of the uniform-prior result of
\S\ref{sec:ablation}: neither the sensor's certified prior nor the actuator's
$\nabla K$-selected support is what the controller runs on.  What it runs on
is measured next.

\begin{table}[htbp]\centering
\small
% generated by codes/aux/gen_gradgate_table.py -- do not edit by hand
\begin{tabular}{lccccc}
\toprule
support $q$ & $1\%$ & $3\%$ & $10\%$ & $30\%$ & $100\%$ \\
\midrule
$\nabla K$ top-$q$ & 16,900 (5/6) & 15,772 (11/12) & 12,166 (6/6) & 8,625 (6/6) & 7,875 (6/6) \\
random-$q$ & 16,687 (4/6) & 17,375 (8/12) & 12,208 (6/6) & 9,541 (6/6) & \multicolumn{1}{c}{n/a} \\
\midrule
no kick & \multicolumn{5}{c}{18,000 (5/6)} \\
\bottomrule
\end{tabular}

\caption{Sparse algorithmic perturbations (modular addition, $p{=}31$,
fraction $0.4$, $20{,}000$-step budget; six seeds per cell, the $3\%$ cells
extended to twelve): mean grok step (seeds grokked) when the kick's
$K$-gradient is restricted to a support of fraction $q$, selected by
$|\partial \KSolsoft/\partial\theta_i|$ (top row) or at random (middle).
Acceleration tracks the support size; the $\nabla K$ selection matters only
in the sparse regime, where it buys reliability.}
\label{tab:gradgate}
\end{table}

\subsection{The measured response to the Occam field}

\paragraph{The response to the Occam field is nucleation-like.}  No linear
regime is resolved over the amplitudes probed.  We apply the field directly rather than
inferring its effect from the controller: from
a snapshot at time $t$ of a plain trajectory, run $300$ steps with and without
a constant pressure $\lambda$ on $\KSolsoft$ (the ``Occam field''), read the
released complexity
$\Delta K(t,\lambda)$, and restore the snapshot (the plain branch is
deterministic, so the $\lambda{=}0$ control is exact).
Fig.~\ref{fig:grokchi}a shows $\Delta K(\lambda)$ at three times along a
memorised trajectory: at $t{=}3{,}000$ the probe releases $2.7$ bits at
$\lambda{=}10^{-8}$, $69$ at $10^{-7}$, and saturates near $2{,}000$ bits
above $10^{-5}$, a threshold-and-saturation curve with no window in which
$\Delta K\propto\lambda$.  By $t{=}13{,}000$ the response is already saturated
at $\lambda{=}10^{-7}$ ($\approx3{,}800$ bits, the map's entire excess over the
true value), with $3{,}600$ released at $3\times10^{-8}$.  These measurements
therefore do not support substituting a fluctuation--dissipation surrogate for
$\chi$: over the tested range, the response is inconsistent with linear
response and instead resembles nucleation---below threshold almost nothing;
above it the basin breaks and the descent self-accelerates to completion.

\paragraph{The finite-field response grows by orders of magnitude toward the
transition.}  The optimal kick time rides its plateau.  Fixing a probe in the onset
regime ($\lambda{=}10^{-7}$, $300$ steps) defines the operational finite-field
response coefficient $\chi(t)=\Delta K/\lambda$.  This is a finite ratio, not
the zero-field derivative of linear-response theory.  We measure it every $500$ steps along
six plain trajectories (Fig.~\ref{fig:grokchi}b).  The released complexity
climbs from order $1$--$40$ bits at $t{=}1{,}000$ (seed-dependent) to
$3{,}800$--$4{,}200$ bits at a peak or plateau.  The location of that peak is
likewise seed-dependent: $t\approx9{,}000$ for the seed that groks
spontaneously at $15{,}250$, as late as $t\approx16{,}000$ for the slowest.  It
always arrives before the visible test-accuracy rise (test $0.09$--$0.15$ at
every seed's peak), and falls once the transition completes, the excess
complexity having been spent.  Since the probe saturates
near the transition, the measured growth is a lower bound.

For the operational test (Fig.~\ref{fig:grokchi}c), we fire a single kick at
time $T$ and measure the delay from kick start to grok over the same six seeds.  The
delay falls broadly with $T$ and collapses to a single check interval ($250$
steps) once $T$ reaches that seed's $\chi$ plateau: five of six seeds
collapse within the sweep, and the sixth is exactly the seed whose plateau
arrives last ($t\approx16{,}000$), its delay down to $500$ at the sweep's
edge.  Fired there, the kick tips the transition essentially instantly.
Kicks fired early, where $\chi$ is orders of magnitude smaller, precondition
the trajectory but must wait thousands of steps for the yield, and on two
seeds a mid-trajectory single kick fails outright within the budget: a lone
pulse with no re-fire can strand the trajectory, which is precisely the
failure mode the staircase's stall rule exists to catch.  Writing $T^\ast$ for
the start time that minimises the total step count to grok (kick start
plus delay, the quantity a practitioner would optimise), $T^\ast$ lies between
$8{,}000$ and $14{,}000$ and sits at the front edge of each seed's plateau: the
seed with the earliest $\chi$ peak has the earliest $T^\ast$ and the seed with
the latest peak the latest, and across the six seeds $T^\ast$ tracks the
$\chi$-peak time at Spearman $\rho=0.84$ ($p=0.04$), as the finite-field-response
picture predicts.

\paragraph{The staircase, explained by the measured response.}  These two measurements account for the
empirically tuned schedule.  (i) The observed nucleation-like response provides
no measured linear regime from which to design smooth annealing:
minimal-dissipation (thermodynamic-length) protocols presuppose a
linear-response metric, which the present measurements do not establish.  For a threshold response
under the standing constraint that the data must stay fit (a sustained
pressure collapses or freezes, \S\ref{sec:controller}), the natural protocol
is bang--bang: full pulses separated by releases, which is the staircase's
shape.  (ii) The pulse should fire where $\chi$ is large, but the plateau's
location is seed-dependent and not observable in advance; the staircase's
stall rule (re-fire when $\KSol$ stagnates above its best) serves as a
causal estimate of ``the trajectory is stuck and responsive'', and its release
rule (a relative fall in $\KSol$) detects that the yield has been realised.
(iii) Iterating dominates waiting: the full staircase reaches
$6{,}500$--$12{,}000$ steps across the six seeds (mean $7{,}875$),
faster on every seed than that seed's best single kick at any $T$
($9{,}250$--$14{,}500$).  Because each pulse lowers the map's complexity, one
possible explanation is that it also raises the response available to
the next pulse: the staircase may build the plateau rather than wait for it.
We have not, however, measured $\chi$ along a kicked trajectory.  The schedule
that was tuned by hand in \S\ref{sec:controller} is thus consistent with a
measured nucleation-like response whose finite-field coefficient is
seed-dependent.

\subsection{The kick's amplitude}

\begin{figure}[htbp]\centering
\includegraphics[width=0.99\textwidth]{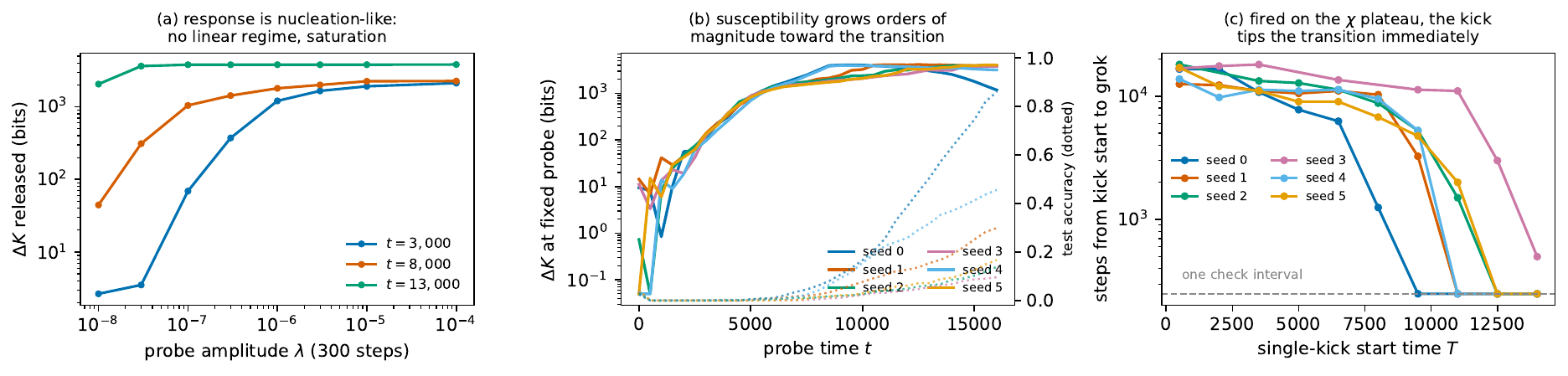}
\caption{Finite-field algorithmic response, measured by direct perturbation (modular
addition, $p{=}31$, fraction $0.4$).  \emph{(a)} Complexity released by a
$300$-step probe of amplitude $\lambda$, at three times along one memorised
trajectory: threshold and saturation, with no linear regime resolved over the
probed amplitudes.  The response is nucleation-like, so these measurements do
not support a fluctuation--dissipation surrogate for $\chi$.  \emph{(b)} The same probe at fixed
$\lambda{=}10^{-7}$ swept along six seeds' trajectories (solid, log scale;
dotted: test accuracy): $\chi$ grows by orders of magnitude and
peaks/plateaus just before each seed's spontaneous transition.
\emph{(c)} Single kicks fired at time $T$, same six seeds: the delay from
kick start to grok collapses to one check interval once $T$ reaches that
seed's $\chi$ plateau (missing points: the kick failed to grok within the
budget; a lone pulse can strand the trajectory, which the staircase's
re-fire rule catches).  The staircase (iterated pulses) beats every seed's
best single-$T$ kick, consistent with, but not proving, the hypothesis that
earlier pulses increase the response available to later ones.}
\label{fig:grokchi}
\end{figure}

\paragraph{The Occam boundary is a misalignment boundary, not an
under-actuation one.}  The finite-size-scaling
picture suggested a companion prediction: the minimum effective kick
amplitude on multiplication should vanish as $f\to f_c^{+}$.  Measured, the
question turns out to be ill-posed.  Comparing $\beta_{\max}$ three decades
apart ($10^{-8}$ and $10^{-5}$, against $\beta{=}0$) at $f\in[0.55,0.8]$, two
seeds each:
every amplitude recovers the true table.  The reason is that at these fractions
the unkicked dynamics already groks in $1{,}250$--$4{,}750$ steps; at
$\beta_{\max}{=}10^{-8}$ the kicked trajectory is check-for-check identical to
$\beta{=}0$ at every fraction and seed.  Larger amplitudes only slow it,
monotonically: at $f{=}0.55$ the two-seed mean rises from $4{,}000$ steps plain
to $6{,}625$ at $\beta_{\max}{=}10^{-5}$, consistent with the data$\times$time
sweep's kick penalty.  Above $f_c$
there is no metastable memorised plateau on multiplication for a kick to tip,
so $\beta_{\min}$ is trivially zero; below $f_c$ no amplitude recovers
(the kick over-simplifies, \S\ref{sec:controller}): the failure is
misalignment of the razor, not insufficient actuation.  Amplitude thresholds
exist only where there is a basin to escape: on \emph{addition} at scarce
data, where the response function of Fig.~\ref{fig:grokchi}a shows a
time-dependent threshold that falls to nothing as the trajectory approaches
its own transition.

\subsection{Matched state perturbations and independent transition diagnostics}
\label{sec:followup}

\begin{figure}[htbp]
\centering
\includegraphics[width=0.99\textwidth]{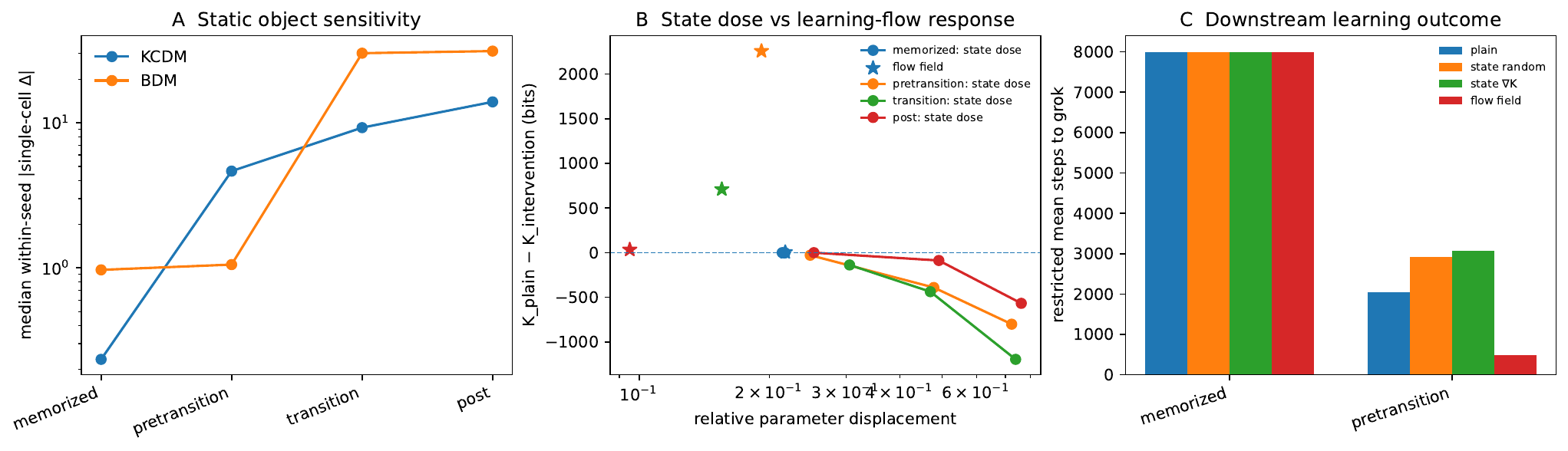}
\caption{Matched object, parameter-state, and learning-flow perturbations
(modular addition, $p{=}31$, fraction $0.4$). \emph{(A)} Median within-seed
absolute response to direct one-cell perturbations of the represented map,
measured independently by $\KSol$ and BDM. \emph{(B)} Random state-reset
dose--response plotted against the observed relative parameter displacement;
stars show the $300$-step Occam-field branch. The ordinate is
$K_{\rm plain}-K_{\rm intervention}$, such that positive values denote greater
complexity release than the matched plain branch. \emph{(C)} Right-censoring–adjusted restricted mean time to grok (RMST) from memorised and pretransition snapshots. The
pretransition Occam-field branch is faster despite a smaller parameter
displacement than either $3\%$ state reset.}
\label{fig:followupflow}
\end{figure}

\paragraph{The pretransition response belongs to the learning flow, not merely
state displacement.}
The original formulation deliberately distinguished classical state
perturbation from perturbing the training flow. We tested that distinction
from identical model--optimiser snapshots. At the pretransition snapshot, a
$300$-step Occam field at $\lambda=10^{-7}$ released a median $2{,}258$ bits of
$\KSol$ (bootstrap $95\%$ CI $1{,}897$--$2{,}747$) and $4{,}159$ BDM bits
($3{,}761$--$5{,}236$), while a random $3\%$ reset moved the corresponding
quantities in the opposite direction ($-390$ and $-626$ bits). The
$\nabla K$-selected $3\%$ reset was still more disruptive ($-749$ and $-868$
bits). This separation was not explained by a larger move in weight space:
the median relative parameter displacement was $0.192$ for the flow field,
versus $0.478$ and $0.576$ for random and $\nabla K$-selected resets. The
flow-versus-random contrasts in $\KSol$, BDM, held-out accuracy, and parameter
displacement were all significant under the paired exact sign-flip test
($p=9.8\times10^{-4}$; $n=11$ pretransition pairs). Static single-cell map
perturbations showed a different phase profile again, with local $|\Delta K|$
and $|\Delta\mathrm{BDM}|$ increasing through the transition rather than
reproducing the pretransition flow peak (Fig.~\ref{fig:followupflow}; full
values in Table~\ref{tab:suppflow}).

The difference also propagated to learning outcome. Starting from the same
pretransition snapshots and counting the common $300$-step actuation window,
all $11/11$ available seeds eventually grokked in every arm, but the
censoring-aware restricted mean time to grok was $482$ steps after the Occam
field, compared with $2{,}050$ for plain continuation and $2{,}914$ after the
random state reset. The flow field shortened restricted mean time by
$1{,}568$ steps versus plain and $2{,}432$ versus random reset (paired
$p=0.00195$ for each). In contrast, when interventions were applied at the
first memorised snapshot, no arm grokked within the $8{,}000$-step follow-up
horizon ($0/12$ each). Thus the intervention is not generically beneficial:
it is sharply phase dependent, consistent with the finite-field susceptibility
measured in Fig.~\ref{fig:grokchi}.

\begin{figure}[htbp]
\centering
\includegraphics[width=0.99\textwidth]{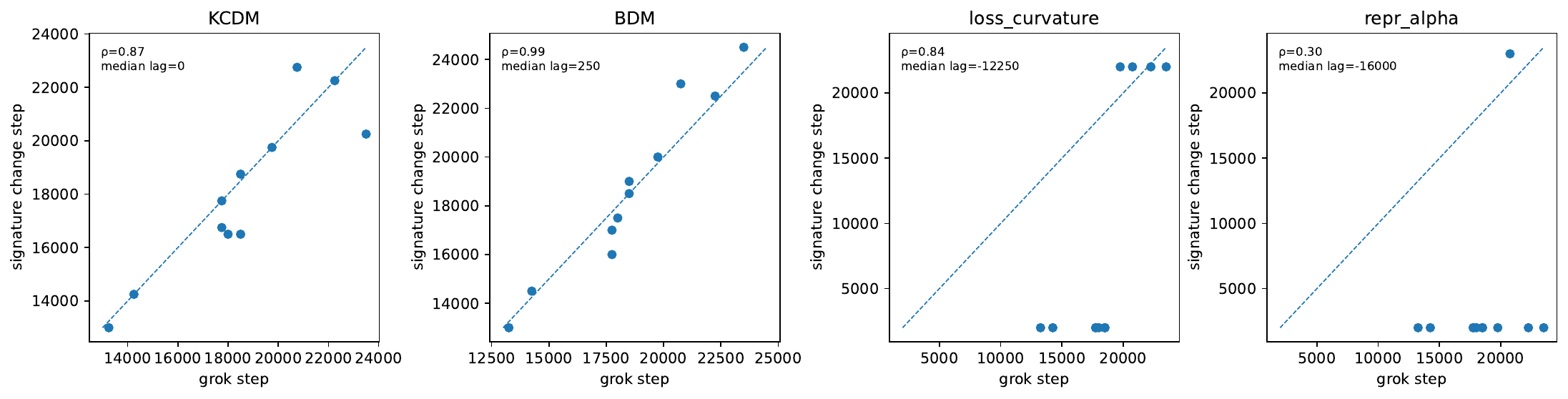}
\caption{Seed-wise timing of candidate transition signatures relative to held-out
grokking ($n=11$ grokking seeds). Dashed lines denote equality between the
signature change point and the grok step. Both $\KSol$ and independently
computed BDM lie near the equality line and strongly preserve the ordering of
seed-specific transition times; loss-curvature and representation-spectrum
changes occur substantially earlier. Exact permutation tests and FDR-adjusted
values are reported in Table~\ref{tab:followuptiming}.}
\label{fig:followuptiming}
\end{figure}

\paragraph{Discrete BDM independently localises transition completion.}
We next asked whether the timing result depended specifically on the
certified differentiable code. For each grokking seed, a change point was
estimated after memorisation from the largest robust one-step change of each
candidate signal and compared with the held-out grok step. The $\KSol$ change
point occurred at a median lag of $0$ steps (95\% CI $-1{,}500$--$0$) and
tracked seed-to-seed grok timing with Spearman $\rho=0.872$ (permutation
$p=7\times10^{-4}$, FDR $q=0.0014$). Independently computed BDM was still
tighter: median lag $250$ steps (one evaluation interval), $\rho=0.986$,
$p=5\times10^{-5}$ and $q=2\times10^{-4}$. Loss curvature also covaried with
the eventual seed-specific transition time ($\rho=0.841$, $q=0.0081$), but its
largest change occurred much earlier (median lag $-12{,}250$ steps), while the
representation tail exponent did not significantly localise completion
($\rho=0.301$, $q=0.451$; Fig.~\ref{fig:followuptiming} and
Table~\ref{tab:followuptiming}). The result therefore separates an early
geometric reorganisation from the algorithmic completion signal used by the
controller.

\begin{table}[htbp]
\centering
\small
\begin{tabular}{lccccc}
\toprule
Signature & $n$ & Median lag to grok (steps) & Spearman $\rho$ & Permutation $p$ & FDR $q$ \\
\midrule
Train fit & 11 & $-18{,}250$ [$-20{,}500$, $-17{,}500$] & -- & -- & -- \\
$\KSol$ & 11 & $0$ [$-1{,}500$, $0$] & 0.872 & $7.0\times10^{-4}$ & 0.0014 \\
BDM & 11 & $250$ [$-500$, $500$] & 0.986 & $5.0\times10^{-5}$ & 0.0002 \\
Loss curvature & 11 & $-12{,}250$ [$-16{,}000$, $-250$] & 0.841 & 0.0061 & 0.0081 \\
Representation $\alpha$ & 11 & $-16{,}000$ [$-17{,}750$, $-12{,}250$] & 0.301 & 0.451 & 0.451 \\
\bottomrule
\end{tabular}

\caption{Timing of memorisation and candidate phase-transition signatures relative
to held-out grokking. Lags are signature step minus grok step and are reported
as median [bootstrap 95\% CI]. Spearman $p$ values are based on $20{,}000$
permutations and adjusted by Benjamini--Hochberg FDR across the four candidate
change-point signals.}
\label{tab:followuptiming}
\end{table}

\paragraph{The coupled collapse requires structured generalisation.}
As a negative control, the same architectures and seed-specific phase times
were trained on fixed shuffled labels. Held-out accuracy remained near chance
and map complexity did not collapse (Fig.~\ref{fig:suppshuffled}). At the
real trajectory's transition phase, structured runs had median $\KSol=1{,}564$
versus $4{,}819$ bits under shuffled labels and median BDM $2{,}558$ versus
$7{,}346$ bits (paired sign-flip $p=0.0156$ for each; $n=7$ matched spectral
pairs). Representation effective rank also separated the conditions
($15.36$ versus $20.02$, $p=0.0156$). Hessian and local-curvature summaries,
however, were not uniformly different across phases
(Figs.~\ref{fig:suppdashboard}--\ref{fig:supphessian}); we therefore treat them
as corroborating geometry rather than as alternative complexity order
parameters or evidence for a universal spectral critical point.

\section{Discussion}
\label{sec:discussion}

\paragraph{Timing, not attribution.}  Four independent checks failed to
find the controller's power where the natural reading of ``certified
algorithmic control'' would place it.  The certified prior is not load-bearing (a uniform
base measure yields bit-identical gate decisions); the per-parameter $\nabla
K$ attribution is not load-bearing above a support threshold (random supports
match); the attribution that does exist is diffuse and transient ($77\%$
readout, ${\sim}5\%$ overlap between kicks); and the measure did not act through the one weight-space readout
available to us.  None of the cheaper signals tested replicated the temporal
structure of the algorithmic signal: the train loss can say ``memorised'' but
never ``reorganised'', a fixed step can say neither, and, among these
observables, only the complexity
collapse marks completion: it is what licenses the release that saves $27\%$ of
the intervention budget, what separates dynamics-shaping from
objective-shaping actuators, and, read as a finite-field response, what predicts when a
kick will tip the transition instantly.  For AID this is a lesson about extending
the calculus from objects to dynamics: for objects the calculus classifies
\emph{elements}; \emph{elements}; for dynamics, what it classifies profitably is
\emph{moments}. The matched follow-up in \S\ref{sec:followup} strengthens
that interpretation: larger parameter resets do not reproduce the
pretransition complexity release or acceleration of the Occam field, while an
independent finite-difference estimator (BDM) localises transition completion
to within one evaluation interval. The shuffled-label control further shows
that the coupled complexity collapse is not generic optimiser drift; at the
same time, the mixed Hessian results argue against replacing the algorithmic
order parameter with a broad claim of spectral criticality.

\paragraph{What, then, does certification buy?}  Not the observed gating
advantage: the tested uniform-prior mixture produced the same gate decisions.
It buys the semantics of the readings.  Because $\KSol$ is an exact
prefix-codelength whose relaxation agrees with it on binary corners and whose invariances
are proved \cite{cdm2026}, the statement ``the kicked run landed at the
true-map complexity'' is an equality of description lengths under a specified
prefix code, rather than merely an equality of unconstrained scores.  Likewise,
interpreting Condition~\ref{cond:occam} as an algorithmic Occam condition
depends on a measure with a defensible claim to estimate complexity.
Certification makes the reading a
falsifiable statement about simplicity under that code, and the ablations then tell us which parts of the
apparatus the engineering actually consumes.

\paragraph{Two axes, two scenarios.}  For the nonlinear-dynamics reader the
system exhibits two distinct scenarios in different control parameters.  Along
the data axis the Occam
boundary exhibits a finite-size-scaling pattern: $f_c$ recedes approximately
as $\ln p/p$, the recovery crossover sharpens with system size, and the
normalisation $n_c\sim p\ln p$ is consistent with a coupon-collector scale
(\S\ref{sec:controller}).  Along the \emph{training flow} at fixed
data, the memorised state behaves as a \emph{metastable} state: the response to
the Occam field is threshold-and-saturation, with no linear regime resolved
over the probed amplitudes; escape is nucleation-like and self-accelerating;
and the finite-field response grows by orders
of magnitude as the spontaneous transition approaches
(\S\ref{sec:calculus}).  The controller's form follows from which scenario it
faces: against metastability the natural protocol is bang--bang pulses, not
annealing.  Against misalignment (below $f_c$) no protocol helps, because
the failure is in the target, not the actuation.

\section{Limitations and open directions}
\label{sec:limits}

\begin{itemize}
\item \textbf{Statistics.} The original dissection experiments run $1$--$12$ seeds per
cell, most cells six (per-seed values for the two ablation grids appear in
Appendix~\ref{app:setup}), so the close original comparisons ($3\%$ top-$q$
vs. random and single-$T$ optima) still require wider replication. The matched
follow-up uses twelve trajectory seeds and paired bootstrap/randomisation
uncertainty, with eight seeds allocated to the expensive spectral diagnostics
(seven matched pairs after excluding a non-grokking trajectory at later
phases). These samples support the large within-seed contrasts reported in
\S\ref{sec:followup}, but remain moderate for fine spectral effects. The
follow-up additionally reports restricted mean time to grok so failed runs are
retained at a fixed horizon rather than discarded from a conditional mean.

\item \textbf{From measured $\chi$ to a closed-loop schedule.}
\S\ref{sec:calculus} accounts for the staircase's form (pulses, stall-fired,
release-on-yield) from the measured response, but does not yet optimise its
constants: an online controller that estimates the local response from its own
recent probe history and sets pulse amplitude and window accordingly is the
natural next step.

\item \textbf{A label-free algorithmic safeguard.}  The transformer's post-release
relapse is cured by a small labelled validation gate
(\S\ref{sec:ablation}), the one component that consumes labels beyond the
supervised training set.  Whether a label-free guard (e.g.\ re-fire on a $\KSol$ rise
above the achieved floor) can replace it is untested.

\item \textbf{Scale of the readout.}  All maps here are small enough to
enumerate ($p^2$ cells, $2^{10}$ parity inputs).  For large systems the map
must be subsampled and $\KSol$ read from a patch ensemble; nothing in the
estimator forbids this, but the control results have not yet been reproduced
in that regime.

\item \textbf{Flow versus state.} The matched resets in
\S\ref{sec:followup} address this distinction for modular addition at
$p{=}31$: resetting $3\%$ of parameters, whether randomly or by $|\nabla K|$,
does not reproduce the pretransition complexity release or downstream
acceleration of the Occam-field branch, despite producing a larger parameter
displacement. This resolves the original open test only in this local setting. Other tasks, architectures, reset operators, and state perturbations remain to
be studied before treating the distinction as universal.

\item \textbf{Weight-norm control.}  It converts into a gateable actuator only in a
narrow, non-monotone window ($\rho{=}0.3$ under weak decay), where it confirms
the objective-shaping prediction (\S\ref{sec:ablation}).  Why the response is
non-monotone in $\rho$, and why no setting reaches the full generalisation the
strong-decay runs attain, are untested here; a version acting on the
initialisation scale between restarts, closer to Omnigrok's own, may behave
differently.
\end{itemize}

\section*{Code and data availability}
The complete grokking harness (tasks, models, controllers), the drivers for
every sweep, and cached results (JSON) for the ablation, susceptibility,
scaling, parity, transformer and data\,$\times$\,time experiments accompany the
manuscript, together with the generators that produce every table and figure.
Two sets of reported numbers are produced by drivers that re-run rather than
cache, and reproducing them needs a GPU: the headline timing means of
\S\ref{sec:controller} (\texttt{grok\_timing.py}) and the data-fraction
complexity climb of \S\ref{sec:controller} (\texttt{grok\_climb.py}).
The six tables and
three of the six figures regenerate from the caches without a GPU;
the remaining three figures re-run their trajectories,
and the two structural identities of Table~\ref{tab:gated} (blind-on weight
decay $\equiv$ plain baseline; uniform-prior sensor $\equiv$ $\sF$ gate
decisions, both seed for seed) are asserted by the table generator itself.
The estimator is the \texttt{pycdm} package of the companion paper
\cite{cdm2026}; setting one environment variable
(\texttt{AIDCONTROL\_RAW=1}) reproduces every number on the raw machine $F$
for the invariance checks quoted in \S\ref{sec:controller}.  The
package is archived at \texttt{10.5281/zenodo.0000000} (final DOI to be
disclosed) and released under the MIT licence at the release tagged for this
manuscript; the repository URL is to be disclosed.

\section*{Declaration of competing interest}
The authors declare that they have no known competing financial interests or
personal relationships that could have appeared to influence the work reported
in this paper.

\section*{CRediT authorship contribution statement}
\textbf{Luan Ozelim:} Conceptualization, Methodology, Software,
Validation, Investigation, Data curation, Formal analysis, Visualization,
Writing -- original draft, Writing -- review \& editing.
\textbf{Abicumaran Uthamacumaran:} Methodology, Software, Investigation,Formal analysis, Visualization,
Writing -- original draft, Writing -- review \& editing.
\textbf{Hector Zenil:} Conceptualization, Supervision, Funding
acquisition.  

\section*{Acknowledgements}
The authors gratefully acknowledge the support of Oxford Immune
Algorithmics and of King's College London.

\section*{Use of AI tools disclosure statement}

Generative AI tools were used to assist with language editing, LaTeX formatting, checking derivations, proposing objections to arguments, and organising the material. All mathematical arguments, proofs, results, and interpretations originated and were verified by the authors, who assumes full responsibility for the manuscript.

\appendix
\section{Experimental setup and per-seed results}
\label{app:setup}

\paragraph{Tasks and models.}  Modular arithmetic: inputs $(a,b)\in
\mathbb{Z}_p^2$, label $(a{+}b)\bmod p$ or $(a\cdot b)\bmod p$; the training
set is a uniformly random fraction $f$ of the $p^2$ cells (seeded), the rest
held out.  MLP: two embeddings of width $d{=}128$ concatenated, two hidden
$\mathrm{ReLU}$ layers of width $256$, linear readout to $p$ classes.
Transformer (\S\ref{sec:ablation}): token embeddings for the two operands,
learned positional embeddings, two pre-norm causal blocks (4 heads, width
$128$, GELU MLP of width $512$), readout at the last position.  Sparse
parity: $n{=}10$ input bits, label the XOR of the first $k$; MLP with two
hidden layers of width $256$.  All training is full-batch AdamW,
learning rate $10^{-3}$, weight decay $1.0$ unless a schedule or the
weak-decay probe says otherwise.

\paragraph{Readout and grok criterion.}  $\KSol$ is evaluated on the argmax
output map over all inputs (bit-planes of the $p\times p$ table; a
$32\times32$ field for parity), every $250$ steps ($50$ in the gated
ablation).  ``Grok step'' is the first check at which held-out accuracy
exceeds $0.9$; means are over grokked seeds with the grokked count reported
alongside.

\paragraph{Staircase constants (kick, parity, transformer).}  Fit tolerance
$\mathrm{CE}<0.01$ ($\mathrm{BCE}<0.03$ for parity); pressure ramp
$\beta\!\leftarrow\!\beta+2\times10^{-5}(\varepsilon-\mathrm{CE})$ clipped to
$[0,3\times10^{-4}]$; release when $\KSol$ falls to $0.6$ of the kick-start
value or after $3{,}000$ steps; re-fire after $4$ stalled checks with
$\KSol>1.05\times$ the best achieved.  Gated ablation
(\S\ref{sec:ablation}): fit gate $\mathrm{CE}<0.05$, release at $0.4$ of the
post-memorisation peak, fixed gate from step $500$; Grokfast
$\hat g=g+\lambda\,\mathrm{EMA}_\alpha(g)$ with $\alpha{=}0.9$,
$\lambda{=}2$; weight-decay schedule $0.1/1.0$; weight-norm projection to
$\rho\,\|\theta_0\|$.

\paragraph{Susceptibility probes.}  Branches restart from a full snapshot
(parameters and optimiser moments); each branch runs $300$ steps with
constant pressure $\lambda$ on $\KSolsoft$ ($\lambda{=}0$: plain); the
$\lambda{=}0$ branch is deterministic, so the control is exact.  The
$\chi(t)$ scan probes every $500$ steps at $\lambda{=}10^{-7}$.

\paragraph{Budgets, seed pools, and one reconciliation.}  The headline kick
numbers of \S\ref{sec:controller} come from dedicated timing runs ($8$ and
$6$ seeds, larger step budgets); the ablations of
\S\ref{sec:ablation}--\ref{sec:calculus} use $6$ seeds with
$20{,}000$--$25{,}000$-step
budgets (the $3\%$ cells of Table~\ref{tab:gradgate}: twelve), and the
single-kick sweep uses $6$ seeds with a $22{,}000$-step budget.  Because full-batch
trajectories are chaotically sensitive to float-level reordering, identical
configurations run through different code paths can shift individual grok
steps by $\mathcal{O}(10^3)$ steps without changing any ordering; all
comparisons in this paper are therefore made within one script's code
path (and the two cross-path identities we do rely on are asserted
exactly; see Code availability).  Every experiment runs on one consumer GPU
in minutes per trajectory.

\paragraph{Per-seed grok steps.}  Tables~\ref{tab:gatedseeds}
and~\ref{tab:gradgateseeds} list the per-seed grok steps behind
Tables~\ref{tab:gated} and~\ref{tab:gradgate}.

\begin{table}[htbp]\centering
\small
% generated by codes/aux/gen_ablation_table.py -- do not edit by hand
\begin{tabular}{llcccccc}
\toprule
actuator & gate policy & \multicolumn{6}{c}{grok step, seeds 0--5} \\
\midrule
Grokfast & \emph{none} (blind off) & 14,050 & 19,050 & 20,350 & -- & -- & 18,250 \\
 & \emph{always} (blind on) & 18,850 & 14,850 & 19,900 & 22,350 & 20,550 & 18,300 \\
 & \emph{fixed} step $\ge$ 500 & 15,100 & 19,150 & 19,450 & -- & 20,000 & 18,850 \\
 & \emph{loss} (on once fit) & 14,550 & 18,050 & 20,200 & 22,500 & 18,750 & 18,950 \\
 & $\KSol$ (on once fit, released) & 14,550 & 18,200 & 20,150 & 22,550 & 18,250 & 18,900 \\
 & \quad$\KSol$, uniform-prior sensor & 14,550 & 18,200 & 20,150 & 22,550 & 18,250 & 18,900 \\
\midrule
weight decay & \emph{none} (blind off) & -- & -- & -- & -- & -- & -- \\
 & \emph{always} (blind on) & 14,050 & 19,050 & 20,350 & -- & -- & 18,250 \\
 & \emph{fixed} step $\ge$ 500 & 17,250 & 15,150 & 17,800 & 19,750 & -- & 19,000 \\
 & \emph{loss} (on once fit) & 15,450 & 14,350 & 16,600 & 20,150 & 22,900 & 17,700 \\
 & $\KSol$ (on once fit, released) & -- & 15,050 & 16,850 & -- & 24,000 & -- \\
\bottomrule
\end{tabular}

\caption{Per-seed grok steps for the gated-vs-blind ablation of
Table~\ref{tab:gated} (``--'': did not reach test $0.9$ within the
$25{,}000$-step budget).}
\label{tab:gatedseeds}
\end{table}

\begin{table}[htbp]\centering
\small
% generated by codes/aux/gen_gradgate_table.py -- do not edit by hand
\begin{tabular}{lcccccc}
\toprule
config & \multicolumn{6}{c}{grok step, seeds 0--5} \\
\midrule
no kick & 15,250 & 18,500 & 18,250 & -- & 18,750 & 19,250 \\
full kick & 6,500 & 8,500 & 7,250 & 6,500 & 6,500 & 12,000 \\
$\nabla K$ top-0.01 & 14,250 & 17,250 & 16,250 & 17,500 & -- & 19,250 \\
$\nabla K$ top-0.03 & 15,500 & -- & 14,750 & 17,750 & 16,000 & 15,750 \\
\quad(seeds 6--11) & 13,750 & 13,750 & 16,000 & 20,000 & 14,750 & 15,500 \\
$\nabla K$ top-0.1 & 14,250 & 11,750 & 9,250 & 10,250 & 13,750 & 13,750 \\
$\nabla K$ top-0.3 & 7,750 & 9,750 & 8,250 & 8,500 & 9,000 & 8,500 \\
random-0.01 & 14,750 & -- & -- & 17,250 & 17,750 & 17,000 \\
random-0.03 & 18,750 & 16,000 & 18,750 & 18,000 & -- & 18,250 \\
\quad(seeds 6--11) & -- & 16,250 & -- & 16,000 & 17,000 & -- \\
random-0.1 & 12,250 & 14,000 & 11,500 & 10,750 & 11,750 & 13,000 \\
random-0.3 & 10,250 & 9,250 & 10,000 & 8,500 & 7,750 & 11,500 \\
\bottomrule
\end{tabular}

\caption{Per-seed grok steps for the sparse-$\nabla K$ grid of
Table~\ref{tab:gradgate} ($20{,}000$-step budget).}
\label{tab:gradgateseeds}
\end{table}

\section{Matched perturbation and learning-geometry follow-up}
\label{app:followup}

The follow-up results support our central claims: the algorithmic signal robustly tracks transition completion, the effective control signal is tied to learning flow timing rather than generic state displacement, and independent BDM/shuffled-label controls argue that the observed collapse reflects structured learning rather than some random optimizer dynamics.

\paragraph{Snapshots and perturbations.}
The follow-up preserves the primary data generator, MLP, optimiser, learning
rate, weight decay, and $\KSol$ implementation. Twelve seeds were trained for
$25{,}000$ steps with diagnostics every $1{,}000$ steps, $\KSol$ checks every
$250$ steps, and BDM checks every $500$ steps. The memorised snapshot is the
first check with train accuracy above $0.99$; ``pretransition'' is the stored
snapshot nearest $2{,}000$ steps before the first test-accuracy crossing of
$0.9$; ``transition'' is that crossing; and ``post'' is $2{,}000$ steps later.
One trajectory did not grok within budget, yielding eleven dynamic
pretransition/transition/post comparisons.

Direct object perturbations changed $96$ uniformly sampled cells of the
predicted $p\times p$ map one at a time by $y_{ij}\mapsto(y_{ij}+1)\bmod p$.
For each cell we recomputed discrete $\KSol$ and BDM using \texttt{pybdm} perturbation analysis,
and reduced cell-level responses to one summary per seed before uncertainty
estimation. Parameter state perturbations reset selected weights exactly to
their own seeded initialisation values. Random supports were evaluated at
$q\in\{0.01,0.03,0.10\}$; the primary $q=0.03$ support was also selected by
the largest $|\partial\KSolsoft/\partial\theta_i|$. Dynamic branches then used
the exact $300$-step plain branch ($\lambda=0$) or Occam-field branch
($\lambda=10^{-7}$) already used in the susceptibility experiment. Relative
parameter displacement was measured as
$\|\theta_{\rm end}-\theta_{\rm start}\|_2/\|\theta_{\rm start}\|_2$ rather
than assuming equal intervention strength.

\paragraph{Downstream outcome and null control.}
For memorised and pretransition snapshots, the four dynamic arms (plain,
random reset, $\nabla K$ reset, Occam field) spent the same initial $300$ steps
in their assigned branch and then continued with plain AdamW up to a common
$8{,}000$-step horizon. Restricted mean time to grok (RMST) assigns
non-grokking runs the horizon and therefore avoids conditioning timing on
success. For the structured-learning null, the same train/test indices,
architecture, optimiser, and seed schedule were used after replacing labels by
a fixed seed-specific random map. Diagnostics were sampled at the phase times
of the corresponding structured run.

\paragraph{Loss and representation geometry.}
Eight seeds were allocated to the expensive deep diagnostics. Local geometry
included curvature along the training-gradient direction and a six-direction
finite-difference curvature estimate. Hessian spectra were estimated with
stochastic Lanczos quadrature ($m=24$, four probes) and summarised by the
largest Ritz value, negative spectral mass, effective rank, and a weighted
Hill exponent of the positive tail. The last quantity is an exploratory
spectral-shape statistic, \emph{not} an estimate of Kolmogorov complexity.
Hidden-representation spectra were computed from the covariance eigenvalues of
the final hidden layer and summarised by effective rank and an analogous
positive-tail exponent.

\paragraph{Statistics.}
The seed is the replicate throughout. Reported medians use $4{,}000$-resample
percentile bootstrap $95\%$ intervals. Matched contrasts use two-sided paired
sign-flip randomisation tests (exact at these sample sizes) and paired rank
biserial effect sizes. Transition timing uses the largest robust one-step
slope after three-point median smoothing, Spearman rank correlation with
$20{,}000$ permutation tests, and Benjamini--Hochberg FDR correction. Grokking
probabilities use Wilson intervals; time-to-grok comparisons use RMST at the
fixed $8{,}000$-step horizon.

\setcounter{figure}{0}
\renewcommand{\thefigure}{S\arabic{figure}}
\setcounter{table}{0}
\renewcommand{\thetable}{S\arabic{table}}

\begin{table}[htbp]
\centering
\small
\textbf{A. Static object sensitivity}\par\smallskip
\begin{tabular}{lcc}
\toprule
Phase & Median within-seed $|\Delta\KSol|$ & Median within-seed $|\Delta\mathrm{BDM}|$ \\
\midrule
Memorised & 0.234 [0.193, 0.679] & 0.969 [0.872, 1.009] \\
Pretransition & 4.64 [4.24, 5.80] & 1.05 [0.926, 2.78] \\
Transition & 9.26 [7.46, 13.10] & 30.18 [29.65, 31.15] \\
Post & 13.93 [11.68, 16.18] & 31.20 [31.19, 31.52] \\
\bottomrule
\end{tabular}

\vspace{0.8em}
\textbf{B. Pretransition dynamic counterfactuals}\par\smallskip
\resizebox{\textwidth}{!}{%
\begin{tabular}{lcccccc}
\toprule
Intervention & $n$ & $K_{\rm plain}-K_{\rm int}$ (bits) & $\mathrm{BDM}_{\rm plain}-\mathrm{BDM}_{\rm int}$ & $\Delta$ test vs. plain & Relative $\theta$ displacement & RMST to grok (steps) \\
\midrule
Plain continuation & 11 & 0 & 0 & 0 & -- & 2,050 \\
Random state reset ($q=0.03$) & 11 & $-390$ [$-431$, $-293$] & $-626$ [$-796$, $-482$] & $-0.061$ [$-0.088$, $-0.049$] & 0.478 [0.436, 0.503] & 2,914 \\
$|\nabla K|$ state reset ($q=0.03$) & 11 & $-749$ [$-842$, $-526$] & $-868$ [$-1,440$, $-705$] & $-0.147$ [$-0.166$, $-0.094$] & 0.576 [0.542, 0.601] & 3,073 \\
Occam field ($\lambda=10^{-7}$) & 11 & 2,258 [1,897, 2,747] & 4,159 [3,761, 5,236] & 0.267 [0.201, 0.361] & 0.192 [0.182, 0.216] & 482 \\
\bottomrule
\end{tabular}}

\vspace{0.35em}
\footnotesize Paired exact sign-flip tests for Occam field versus random reset:
$p=9.8\times10^{-4}$ for $\KSol$ release, BDM release, test-accuracy change,
and relative parameter displacement; $p=0.00195$ for RMST. All $11/11$
pretransition seeds grokked in each downstream arm within the $8{,}000$-step
horizon.

\caption{Selected matched object/state/flow results. Panel A reports static
single-cell sensitivity after within-seed reduction. Panel B reports the
pretransition dynamic branches. Dynamic $\Delta K$ and $\Delta\mathrm{BDM}$
are the matched plain-window value minus the intervention value, so positive
numbers indicate complexity release relative to plain continuation. RMST is
measured from the common starting snapshot and includes the $300$-step
actuation window.}
\label{tab:suppflow}
\end{table}

\begin{figure}[htbp]
\centering
\includegraphics[width=0.94\textwidth]{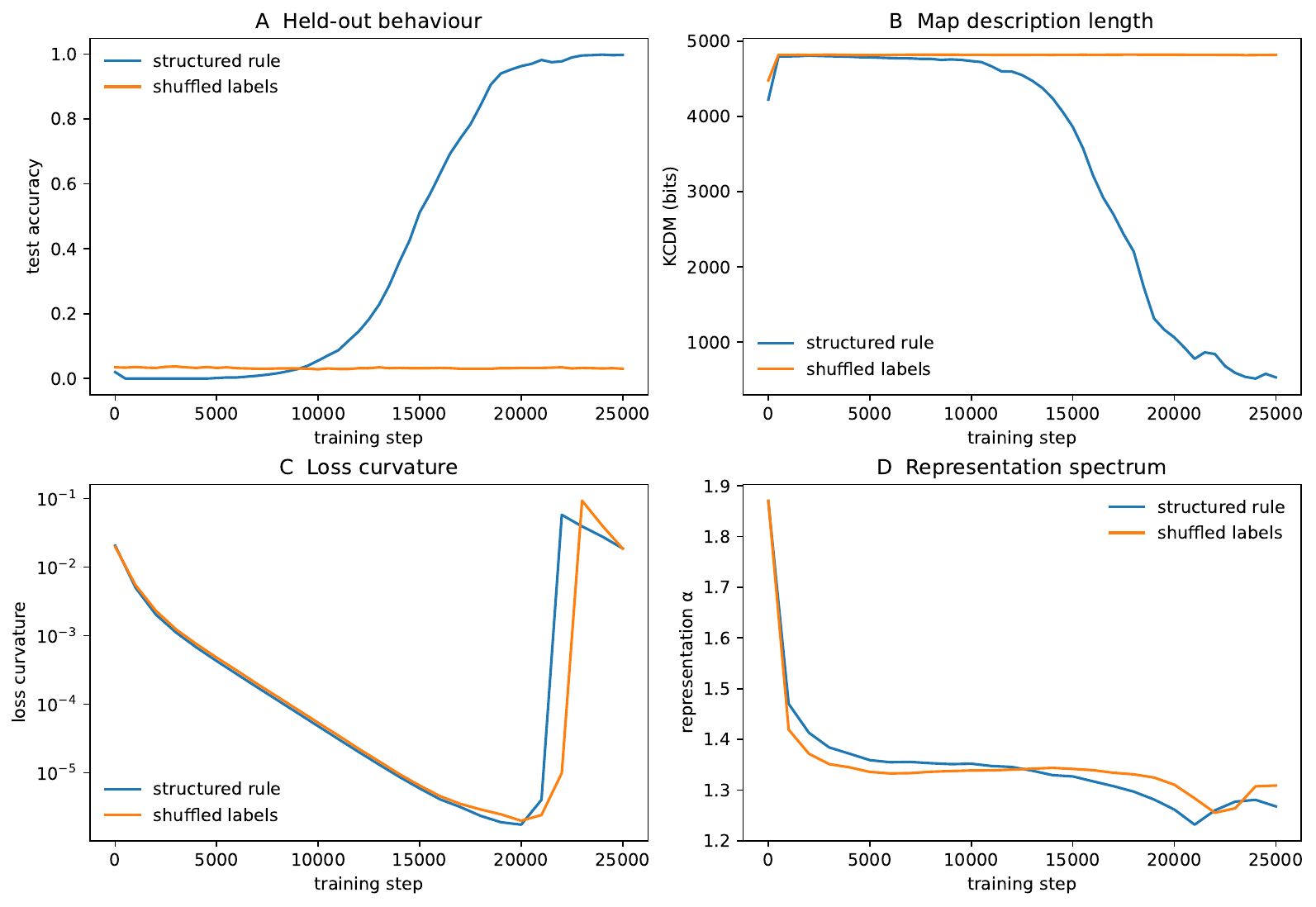}
\caption{Structured modular addition versus the same-seed shuffled-label null.
Held-out generalisation and the collapse in map description length occur only
for the structured rule. Loss-curvature and representation-spectrum changes
are shown as independent geometric measures or diagnostic comparable to complexity estimators.}
\label{fig:suppshuffled}
\end{figure}

\begin{figure}[htbp]
\centering
\includegraphics[width=0.94\textwidth]{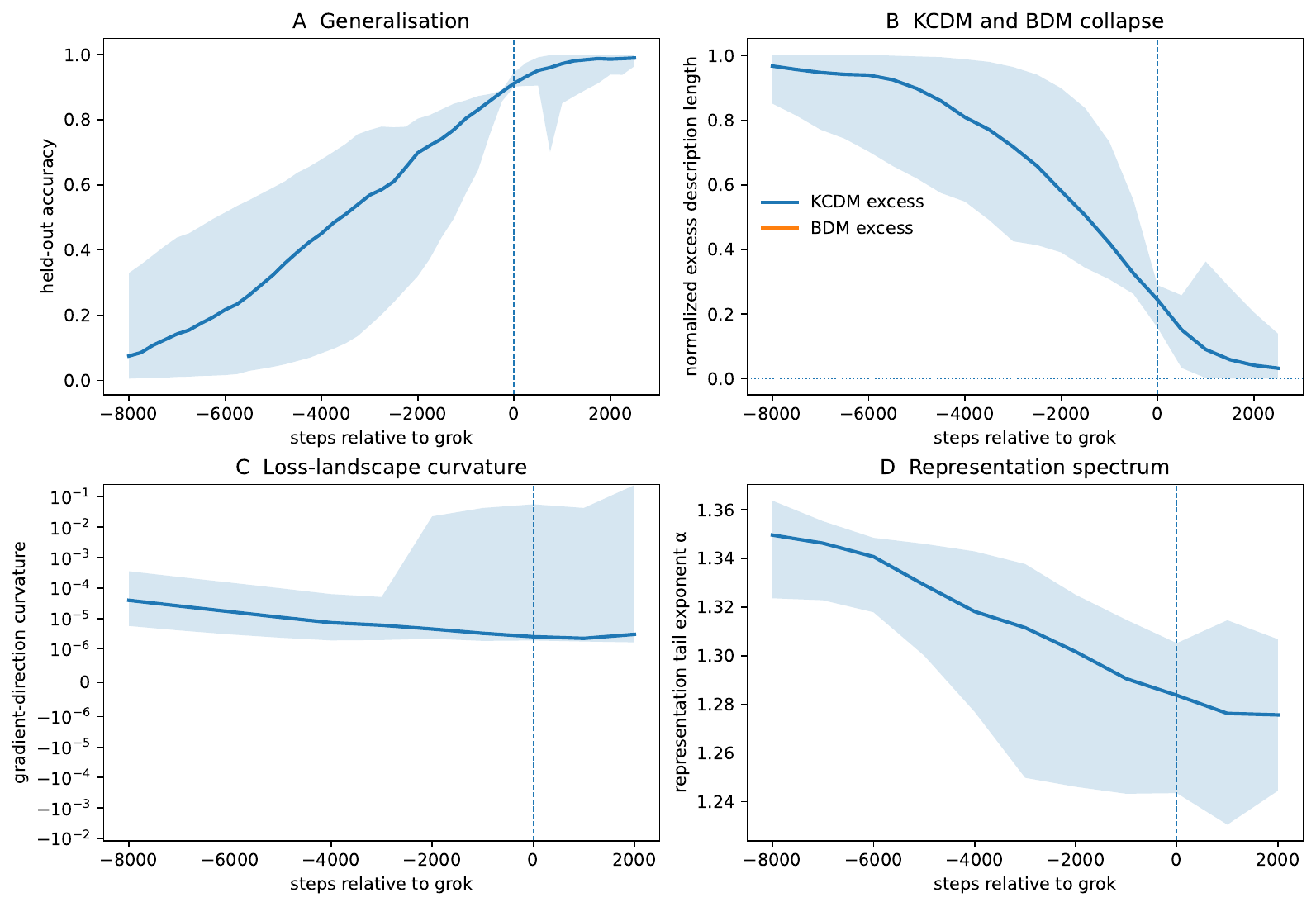}
\caption{Grok-aligned phase-transition signature across structured trajectories. The
held-out accuracy rise is accompanied by a collapse of both $\KSol$ and BDM
excess description length. Loss curvature and the representation tail
exponent evolve earlier and more gradually, consistent with the timing
analysis in Table~\ref{tab:followuptiming}.}
\label{fig:suppdashboard}
\end{figure}

\begin{figure}[htbp]
\centering
\includegraphics[width=0.99\textwidth]{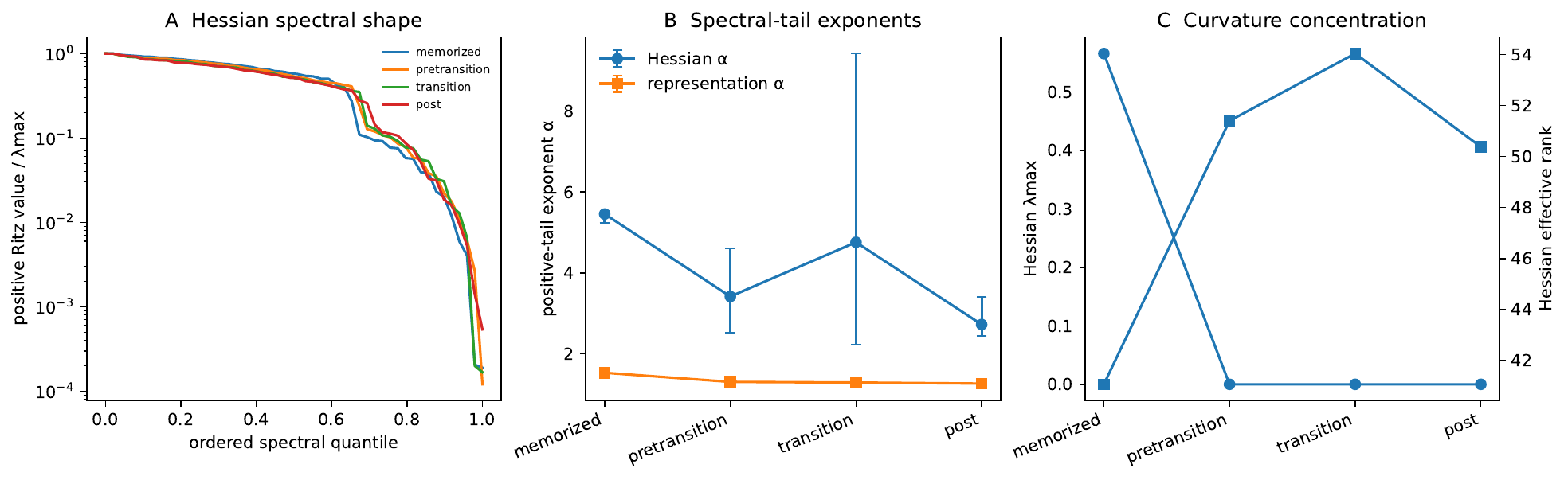}
\caption{Exploratory Hessian and representation-spectrum diagnostics across
phases. Stochastic-Lanczos positive spectral shapes, positive-tail exponents,
$\lambda_{\max}$ and Hessian effective rank change across training, but the
phase-wise shuffled-label tests are mixed; these quantities are therefore
reported as supporting geometry and not as replacements for the algorithmic
order parameter.}
\label{fig:supphessian}
\end{figure}

\end{document}